%% file: arxiv.tex
\documentclass{article}

\usepackage[table]{xcolor}

\usepackage[preprint]{corl_2026} % Uncomment for pre-prints (e.g., arxiv); This is like ``final'', but will remove the CORL footnote.

\usepackage{graphics} % for pdf, bitmapped graphics files
\usepackage{times} % assumes new font selection scheme installed
\usepackage{amsmath} % assumes amsmath package installed
\usepackage{amssymb}  % assumes amsmath package installed
\usepackage{booktabs}

\usepackage{enumitem}
\usepackage{wrapfig}
\usepackage{mathtools}
\usepackage{tabularx}
\usepackage{array}
\usepackage{makecell}
\usepackage{multirow}
\usepackage{multicol}
\usepackage{graphicx}
    \graphicspath{{figure/}}
\usepackage[font={small}]{subcaption}
\usepackage[font={small}]{caption}
\usepackage{algorithm}
\usepackage{algpseudocode}
\usepackage[scaled=0.9]{inconsolata}
\usepackage{ulem}

\usepackage[utf8]{inputenc}
\usepackage[T1]{fontenc}

\usepackage{tcolorbox}
\tcbuselibrary{listings,breakable}
\usepackage{listings}
\definecolor{codegreen}{rgb}{0,0.6,0}
\definecolor{codegray}{rgb}{0.5,0.5,0.5}
\definecolor{codepurple}{rgb}{0.58,0,0.82}
\definecolor{backcolour}{rgb}{0.95,0.95,0.95}
\definecolor{mycitecolor}{RGB}{71, 191, 38}
\definecolor{mylinkcolor}{RGB}{40, 115, 201}
\hypersetup{
  colorlinks=true,
  citecolor=mycitecolor,
  linkcolor=mylinkcolor,
  urlcolor=mycitecolor,
}
\lstdefinestyle{mystyle}{
  backgroundcolor=\color{backcolour},   commentstyle=\color{codegreen},
  keywordstyle=\color{magenta},
  numberstyle=\tiny\color{codegray},
  stringstyle=\color{codepurple},
  basicstyle=\ttfamily\scriptsize,
  breakatwhitespace=true,         
  breaklines=true,
  breakindent=0pt,
  captionpos=t,       
  columns=fullflexible,
  keepspaces=true,        
  frame=single,
  showspaces=false,                
  showstringspaces=false,
  showtabs=false,                  
  tabsize=2
}
\newtcolorbox{promptbox}[1][]{
  enhanced,
  breakable,
  colback=backcolour,
  colframe=black,
  boxrule=0.4pt,
  arc=0pt,
  left=4pt,
  right=4pt,
  top=4pt,
  bottom=4pt,
  listing only,
  listing options={style=mystyle,frame=none},
  #1
}

\title{CRAFT: Coaching Reinforcement Learning Autonomously using Foundation Models for Multi-Robot Coordination Tasks}

\author{
  Seoyeon Choi\\
  Department of Mechanical Engineering\\
  University of California Berkeley 
  United States\\
  \texttt{seoyeon99@berkeley.edu} \\
  \And
  Kanghyun Ryu \\
  Department of Mechanical Engineering\\
  University of California Berkeley 
  United States\\
  \texttt{kanghyun.ryu@berkeley.edu} \\
  \AND
  Jonghoon Ock \\
  Department of Mechanical Engineering\\
  University of California Berkeley 
  United States\\
  \texttt{ockjh228@berkeley.edu} \\
  \And
  Negar Mehr \\
  Department of Mechanical Engineering\\
  University of California Berkeley 
  United States\\
  \texttt{negar@berkeley.edu} \\
}

\begin{document}
\maketitle
% \vspace{-2em}

%===============================================================================

\begin{abstract}
    % The purpose of this document is to provide both the basic paper template and submission guidelines. Abstracts should be a single paragraph, between 4--6 sentences long, ideally. Gross violations will trigger corrections at the camera-ready phase.
    Multi-Agent Reinforcement Learning (MARL) provides a powerful framework for learning coordination in multi-agent systems. 
    However, applying MARL to robotics remains challenging due to their high-dimensional continuous joint action spaces, complex reward design, and non-stationarity from concurrently learning agents.
    % , and non-stationary transitions inherent to decentralized settings. 
    % On the other hand, humans learn complex coordination through staged curricula, where long-horizon coordination behaviors are progressively built upon simpler skills. 
    On the other hand, humans often learn complex coordination with the help of coaches, who guide learning through carefully designed curricula and detailed feedback. Building on the reasoning capabilities of foundation models, we argue that these models can similarly \textit{coach} robots to learn coordination.
    Motivated by this, we propose CRAFT: \underline{C}oaching \underline{R}einforcement learning \underline{A}utonomously using \underline{F}oundation models for learning coordination \underline{T}asks, a framework that leverages foundation models to act as a ``coach'' for multi-robot coordination. CRAFT automatically decomposes long-horizon coordination tasks into sequences of subtasks using the planning capability of Large Language Models (LLMs). Then, CRAFT trains each subtask using LLM-generated reward functions, and refines them through a Vision Language Model (VLM)-guided reward-refinement loop. We evaluate CRAFT on multi-quadruped navigation and bimanual manipulation tasks, and demonstrate its capability to learn complex coordination behaviors. In addition, in a multi-quadruped navigation setting, we show that our learned policies transfer to the real world.
    % In addition, we validate the multi-quadruped navigation policy in real hardware experiments. 
    Project website is \href{https://iconlab.negarmehr.com/CRAFT/}{https://iconlab.negarmehr.com/CRAFT/}
\end{abstract}

 % Curriculum Learning. 

%===============================================================================

\section{Introduction}

Learning coordination behaviors is essential for enabling multiple robots to work together in real-world tasks. In practical robotic systems, coordination should often be achieved under decentralized execution, where each agent makes decisions without access to other agents' global states, actions, or intentions~\citep{busoniu2008comprehensive}, such as multiple quadrupeds navigating a cluttered environment while avoiding collisions without explicit knowledge of other agents' future motions. While this enables scalable and reliable deployment, it also makes coordination challenging: agents must reason about other's behavior through interaction alone. Multi-Agent Reinforcement Learning (MARL) provides a promising framework for such decentralized decision-making, with notable successes in real-time strategy games~\citep{starcraft,alphastar} and simulated sports environments~\citep{googlefootball}, and is increasingly being explored for multi-robot applications~\citep{marladona,multi-quadruped-soccer}. However, applying MARL to real-world robotics remains difficult due to high-dimensional action spaces inherent in many multi-agent tasks, sparse-reward nature of coordination, and the decentralized nature of policies. These challenges are further amplified in physically interactive and long-horizon tasks, where agents must coordinate contact-rich behaviors over multiple stages and small errors in early interactions can prevent later task success.

On the other hand, humans often learn complex coordination with guidance from a coach. For example, a soccer coach designs training drills that progressively build team coordination, provides learning signals to guide the teammates, and evaluates whether the team is ready to advance or should revise its current training. We argue that, similarly, in MARL, a coaching process that structures training, provides learning signals, and evaluates progress can help agents acquire fundamental skills before building toward complex collaborative behaviors that are difficult to learn through direct training~\citep{marladona,manual-n-agent-curriculum,portelas2021automatic}. However, designing such guidance remains nontrivial, as it requires task-level reasoning to identify useful intermediate subtasks, define appropriate learning signals, and evaluate learning progress. Recent advances in foundation models, such as Large Language Models (LLMs) and Vision-Language Models (VLMs), provide a natural way to automate these roles. LLMs can decompose tasks and generate rewards~\citep{progprompt,voyager,eureka,vlm-zeroshot-reward}, while VLMs can evaluate visual rollouts and provide feedback on policy behavior~\citep{aha-vlm-evaluation,autoeval}. We argue that these capabilities can be integrated into a single \emph{coaching scheme} that teaches agents to coordinate by decomposing tasks, defining subtask success criteria, and providing feedback for improvement.

% On the other hand, humans learn complex coordination progressively. For example, a soccer team begins with simple cooperative drills, such as short passes, before advancing to elaborate team strategies. Inspired by this process, curriculum generation has been studied as an effective approach for learning coordination in MARL by structuring training into stages of increasing complexity~\citep{chen2021variational}. By decomposing long-horizon objectives into manageable subtasks, curricula allow agents to first learn fundamental skills and then build on them to acquire complex collaborative behaviors that are difficult to learn by training directly~\citep{manual-n-agent-curriculum,marladona,portelas2021automatic}. However, designing such curricula remains nontrivial, as it requires domain knowledge to identify useful intermediate steps and reasoning ability to monitor learning progress~\citep{wang2021survey}. Notably, recent advances in foundation models, such as Large Language Models (LLMs) and Vision-Language Models (VLMs), have demonstrated emergent reasoning capabilities that have been leveraged for task decomposition~\citep{progprompt,voyager}, reward shaping~\citep{eureka,vlm-zeroshot-reward}, and policy evaluation~\citep{aha-vlm-evaluation,autoeval}. We argue that these capabilities can be integrated into a single \emph{coach}, an entity that teaches agents how to coordinate by decomposing tasks, defining subtask success criteria, and providing feedback to guide improvement.

\begin{figure}
    \centering
    \begin{subfigure}{0.70\linewidth}
        \centering
        \includegraphics[width=\linewidth]{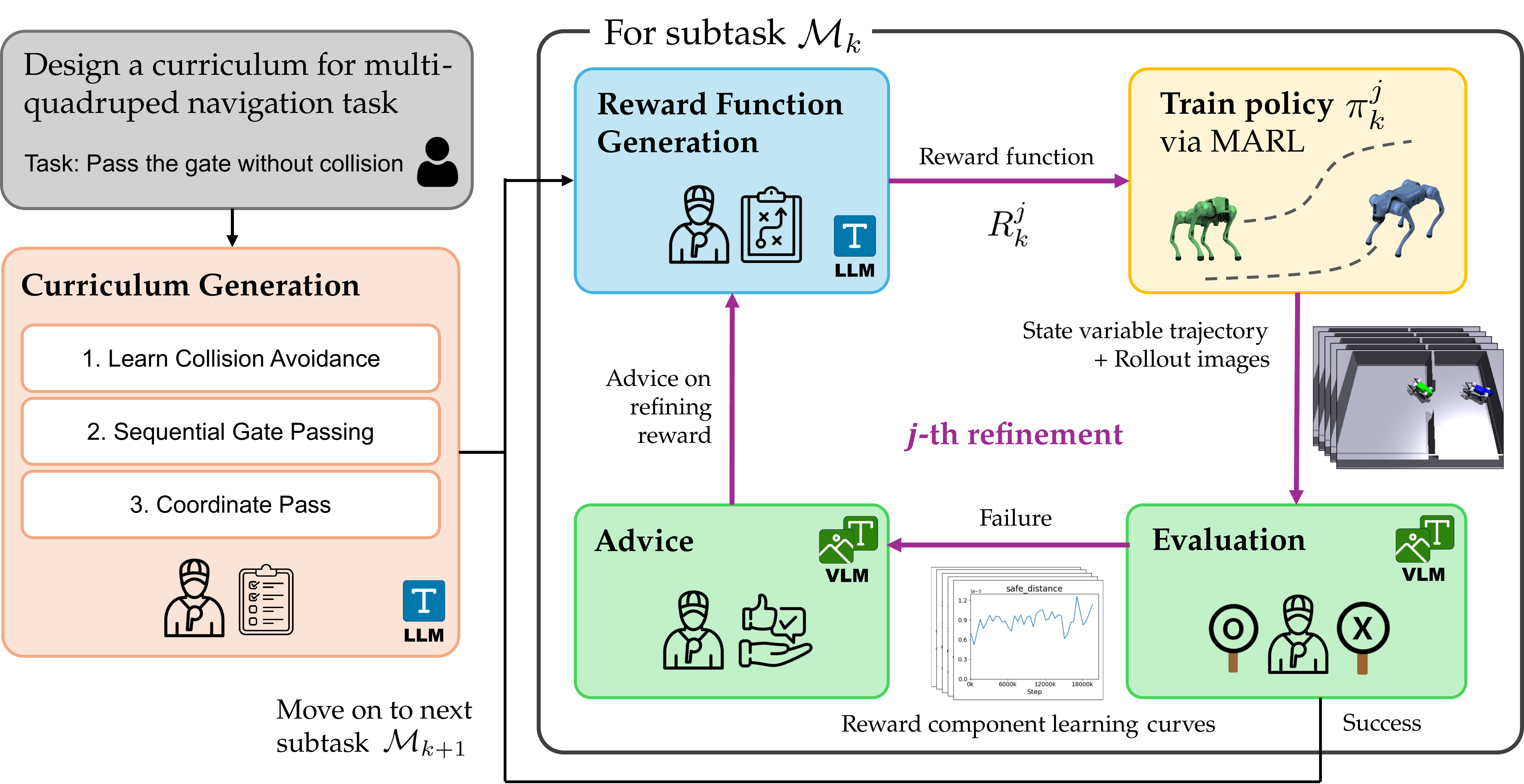}
    \end{subfigure}
    \hfill
    \begin{subfigure}{0.28\linewidth}
        \centering
        \includegraphics[width=\linewidth]{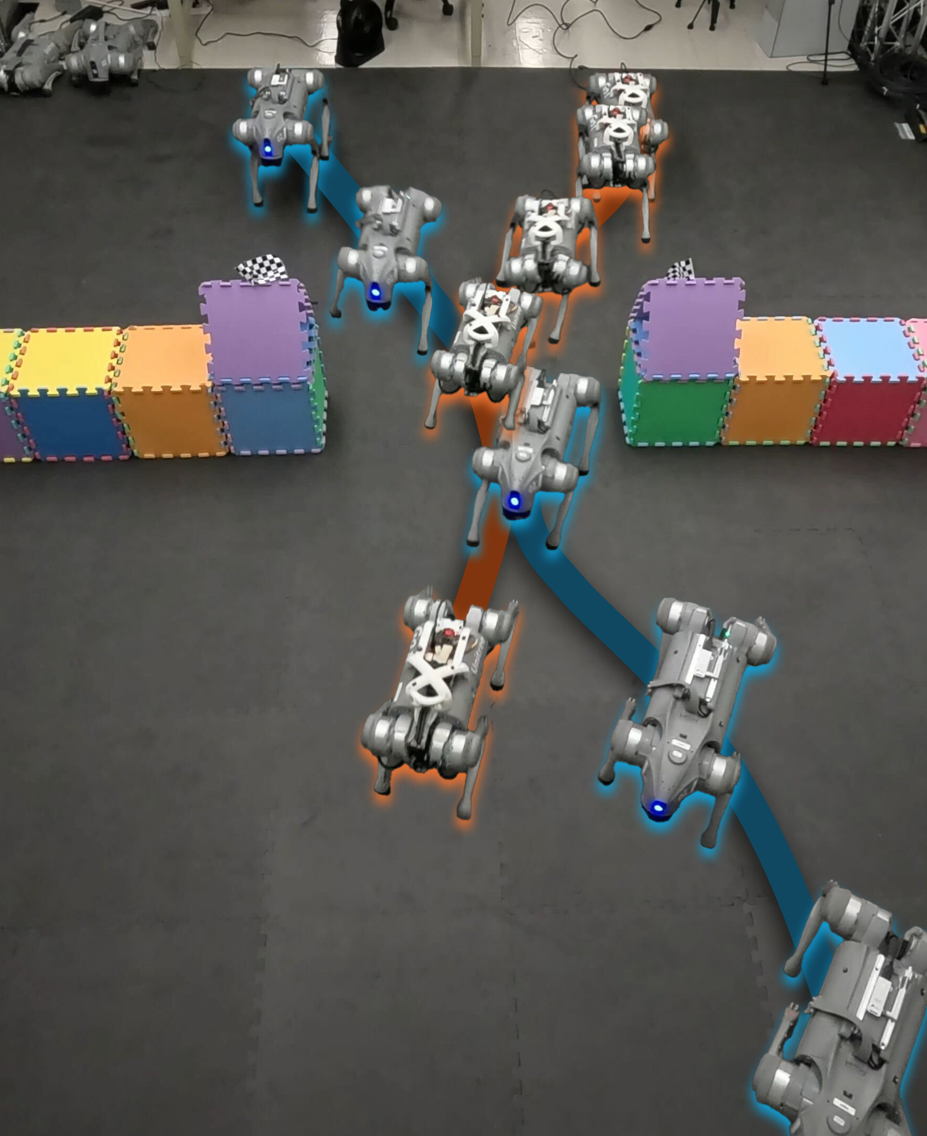}
    \end{subfigure}
    \caption{
    % (Left) Overview of \textbf{CRAFT}. Given a natural language description of the target task, CRAFT generates a curriculum $C = [\mathcal{M}_1, \dots, \mathcal{M}_K]$ of subtasks that progressively build coordination towards the target task. For each subtask $\mathcal{M}_k$, a reward LLM generates a reward function $R^j_k$ and a decentralized policy $\pi^j_k$, shared among agents, is trained via MARL. A VLM evaluates this policy using state variable trajectories and snapshot images. If the policy fails, an advice VLM provides advice from reward component learning curves, plots of individual reward terms during training. The reward LLM incorporates this advice to generate a revised reward. Successful policy advances to the next subtask and this iterative process continues until the final task is reached. 
    % (Right) Demonstration of the Quadruped Gate task, where two quadrupeds coordinate to pass through a narrow gate without collision: the blue robot passes first while the orange robot waits; then, the orange robot proceeds once the gate is clear. This policy was learned in simulation and transferred to real-world.
    (Left) Overview of \textbf{CRAFT}. A curriculum LLM decomposes the target multi-agent task into a sequence of subtasks that progressively build towards coordination. For each subtask $\mathcal{M}_k$, a reward LLM generates a reward $R_k^j$, MARL trains a decentralized policy $\pi_k^j$, and a VLM evaluates rollout images and state trajectories. If the policy fails, an advice VLM generates advice on refining the reward from reward component learning curves. Otherwise, the curriculum proceeds to the next subtask. 
    (Right) Demonstration of the Quadruped Gate task, where two quadrupeds coordinate to pass through a narrow gate without collision.
    }
    \label{fig:overall_method}
\end{figure}

In this paper, we introduce \textbf{CRAFT}: \textbf{C}oaching \textbf{R}ein\-forcement learning \textbf{A}utonomously using \textbf{F}oundation models for multi-robot coordination \textbf{T}asks.
% , a framework that leverages foundation models as a coach that automatically generates curricula for long-horizon multi-robot coordination tasks. 
% Using the curricula design, reward generation, and policy evaluation from foundation models, CRAFT can learn complex coordination tasks that cannot be achieved without precisely tuned reward functions or other human expert intervention.  
In CRAFT, 
% analogous to a human coach who assigns progressively challenging drills to develop complex team strategies, 
our foundation model coach first decomposes a complex cooperative task into a sequence of subtasks that facilitate the learning of multi-agent coordination policies. To guide each subtask, an LLM generates an executable reward function, while a VLM evaluates trained policies and provides feedback to refine the LLM-generated reward through a VLM-guided reward-refinement loop. Through this process, our framework CRAFT produces curricula for multi-agent coordination tasks, designs semantically rich reward functions, and evaluates task-policy alignment using visual information. Our overall framework is shown in Fig.~\ref{fig:overall_method}. Through experiments in simulation and on physical robots, we demonstrate that CRAFT enables learning coordination behaviors 
% in multi-quadruped navigation and bimanual manipulation tasks 
in a number of tasks that standard learning methods fail to solve.

In summary, our contributions are the following: 1) We propose \textbf{CRAFT}, a framework that automatically coaches multi-robot teams for long-horizon coordination tasks by leveraging the reasoning capabilities of foundation models;
2) We evaluate CRAFT on multi-agent coordination tasks in multi-quadruped navigation and bimanual manipulation tasks and demonstrate learning complex coordination policies that were difficult to learn without proper curricula or reward design;
3) We show that the learned policies transfer to hardware in a multi-quadruped navigation task, demonstrating their real-world transferability.
% We validate the multi-quadruped navigation policies learned with CRAFT in hardware experiments, demonstrating their real-world transferability.

\section{Related Works}

% \subsection{Foundation Models for Reinforcement Learning}

% \subsection{Curriculum Learning and Foundation Models}
% \vspace{-3pt}

\textbf{Curriculum Learning}. Long-horizon RL tasks require agents to discover complex behaviors through sparse and delayed feedback which makes exploration inefficient and unstable. Curriculum learning addresses this challenge by decomposing training into a sequence of subtasks with increasing difficulty~\cite{portelas2021automatic}. However, designing effective curricula is inherently challenging as it requires decomposing complex tasks into meaningful sequences of subtasks and reasoning about their relative difficulty to ensure a coherent learning progression. As a result, many prior approaches to curriculum design rely on extensive trial-and-error by human experts~\cite{manual-n-agent-curriculum,tizero}, which limits their applicability to diverse environments. To overcome these limitations, more recent efforts have explored leveraging the reasoning capabilities of LLMs to enable automated curriculum design~\cite{eurekaverse,curricullm,aura}. In this work, we extend curricula design with foundation models to multi-agent settings, where reasoning about agent interactions and long-horizon task decomposition is even more critical.

% \subsection{LLM/VLM for Reward Design}
% \vspace{-3pt}
\textbf{LLM/VLM for Reward Design}. Recent works have shown the effectiveness of reasoning and general knowledge of foundation models for reward design. Several works leverage language models to generate reward functions in executable code and further refine them through evolutionary search~\cite{eureka,song2023self}. Others use VLM's ability to evaluate alignment between visual observations and language descriptions, directly querying a VLM for the reward~\cite{vlm-zeroshot-reward} or providing other feedback to the policy such as success or failure~\cite{aha-vlm-evaluation,wiping-curriculum,autoeval}. Building on these works, we employ the reward generation capabilities of LLMs to produce reward functions for each subtask in a curriculum, and use a VLM to evaluate trained policies and provide feedback for refining reward functions.

% \subsection{Multi-robot Coordination}
% \vspace{-3pt}

\textbf{Multi-robot Coordination}. Several works have explored MARL for multi-robot coordination. For a team of quadrupeds, MARL has been employed in a hierarchical structure where it learns high-level coordination behavior for object manipulation~\cite{cable-towed-qudruped,mapush} or soccer~\cite{multi-quadruped-soccer}, combined with low-level locomotion policy. However, these methods rely on either carefully designed rewards~\cite{mapush} or manually specified curricula~\cite{cable-towed-qudruped}. MARL also has been used for multi-robot navigation with collision avoidance in aerial~\cite{CBVF-MARL} or mobile robots~\cite{sharma2024learning}. Notably, curriculum learning frameworks have been actively employed in MARL training~\cite{cable-towed-qudruped,CBVF-MARL,multi-quadruped-soccer} due to their effectiveness in learning complex coordination behaviors. Inspired by the significance of curriculum learning in MARL, our aim is to automate curriculum design using foundation models to enable learning complex coordination behaviors while reducing the need for extensive human intervention in MARL and pave the way for more extensive use of MARL in multi-robot tasks.

\section{Problem Formulation}

In this work, we aim to learn multi-agent coordination policies where a team of agents should coordinate to achieve a target task $\mathcal{M}_{target}$. We model the target task as a decentralized partially observable Markov decision process (Dec-POMDP~\cite{mappo}), defined by the tuple $\mathcal{M}_{target} = \langle N, \mathcal{S}, \mathcal{A}, O, P, R_{target}, \gamma \rangle$, where $N$ is the number of agents, $\mathcal{S}$ is the global state space, and $\mathcal{A}$ is the joint action space. We denote the local observation of each agent $i$ at the global state $s_t$ as $o_t^i = O(s_t; i)$, where subscript $t$ denotes timestep. The joint action of all $N$ agents is denoted as $a_t = (a_t^1, \dots, a_t^N)$, where $a_t^i$ is the action of agent $i$ at time $t$. The transition probability from state $s_t$ to $s_{t+1}$ when taking the joint action $a_t$ is denoted as $P(s_{t+1}|s_t, a_t)$. There exists a shared reward function $R_{target}(s_t, a_t, s_{t+1})$ that agents need to collectively optimize, and $\gamma$ is the discount factor. In this work, we train a MARL policy using a centralized training and decentralized execution (CTDE) framework~\cite{mappo, qmix}, where all agents share the same policy $\pi_\theta (a_t^i|o_t^i)$ that observes each agent's local observation $o_t^i$ and outputs its action $a_t^i$. Our goal is to learn the shared policy $\pi_\theta (a_t^i|o_t^i)$ such that it maximizes the expected discounted reward $J(\theta) = \mathbb{E}_{a_t, s_{t}} \left[\sum_{t}^{\infty} {\gamma^t R_{target}(s_t, a_t, s_{t+1})} \right]$. 

We define a curriculum $\mathcal{C}$ as a sequence of $K$ subtasks $\mathcal{C} = [\mathcal{M}_1, \mathcal{M}_2, \dots, \mathcal{M}_K]$, where $\mathcal{M}_k$ denotes the $k$-th subtask. Each subtask $\mathcal{M}_k$ includes a reward function $R_k$ which quantifies how the subtask is defined. A curriculum can equivalently be described as a sequence of reward functions $\left[R_1, R_2, \dots, R_K \right]$. Our goal is to find a sequence of reward functions $\left[R_1, R_2, \dots, R_K \right]$ such that when the policy is trained sequentially on it, the resulting policy can accomplish $\mathcal{M}_{target}$.

\section{Method}

Motivated by how human coaches train teams for coordination tasks, our method leverages LLMs and VLMs as a coach for teaching robots on how to coordinate. In training a team for complex coordination tasks, a coach would first decompose the target task into a sequence of subtasks that specify the details of the coordination strategy. The coach then trains the team on each subtask by defining its objectives (rewards). Once training is complete, the coach evaluates the team’s performance, deciding whether to progress to the next subtask or revisit the current one. Inspired by this process, our framework follows four key stages as shown in Fig.~\ref{fig:overall_method}:
\begin{enumerate}[leftmargin=*]
    \item \textbf{Curriculum generation module} -- A curriculum LLM decomposes the long-horizon coordination task into a sequence of subtasks described in natural language.
    \item \textbf{Reward function generation module} -- A reward generation LLM generates a reward function in executable python code based on the natural language descriptions of the subtasks, and provides dense rewards that clearly specify the desired behavior for each subtask.
    \item \textbf{Policy evaluation module} -- An evaluation VLM evaluates the success or failure of the policy trained with the LLM-generated reward based on the visual and quantitative rollouts of the policy. 
    \item \textbf{Reward refinement module} -- If the policy fails to achieve the desired behavior for each subtask, an advice VLM provides advice on how to change the reward based on the rollout information and learning curve. Then, an LLM takes the advice and refines the reward function.
    % \item \textbf{Sequential training of subtasks} -- Throughout the training, we initialize each subtask with the policy learned from the previous one while motivating exploration to learn the new subtask.
\end{enumerate}

\noindent In the following, we discuss each of these modules in detail. Full prompts are in Appendix~\ref{sec: prompt}.

\begin{figure*}[!t]
    \centering
    \includegraphics[width=\linewidth]{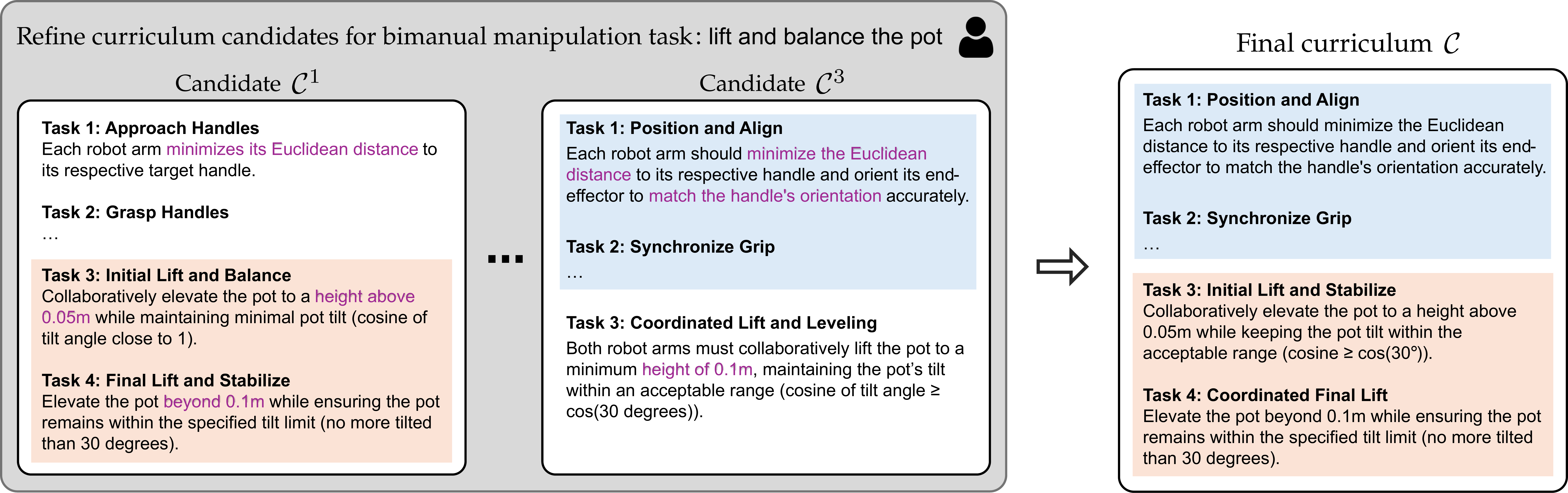}
    \caption{Example of curriculum refinement for task \texttt{lift and balance the pot}. 
    Three different candidate curricula $\mathcal{C}^1$ to $\mathcal{C}^3$, generated by the curriculum module, are re-provided to the LLM for refinement. In $\mathcal{C}^1$, Task 1 focuses only on minimizing distance, while Task 1 in $\mathcal{C}^3$ is defined as minimizing distance and matching orientation. In contrast, Task 3 and Task 4 in $\mathcal{C}^1$ break down the lifting into two stages of first lifting halfway and then to a full height, whereas $\mathcal{C}^3$ represents lifting as a single task. The curriculum LLM merges these candidates into a final curriculum $\mathcal{C}$ by selecting the stronger tasks definitions from each candidate.}
    \label{fig:curriculum}
\end{figure*}

\subsection{Curriculum Generation Module} \label{subsec: curriculum_generation}
First, we break down a coordination task into a sequence of subtasks. Given a natural language description of the target task and environment, a curriculum LLM outputs a sequence of subtasks in natural language. For example, for a bimanual pot-lifting task, the curriculum may guide the arms to approach the pot, grasp it, and lift it together. However, relying solely on zero-shot curriculum generation can result in unstable outputs due to the stochastic nature of LLM responses, such as curricula that may be overly simple, unnecessarily long, or misaligned with the target task.
% \vspace{-3pt}

To mitigate this, we first generate multiple candidate curricula $\mathcal{C}^{i}, i \in \{1,2,\dots,I\}$. We then prompt the LLM again to refine these candidates into a more coherent and effective curriculum. An example is shown in Fig.~\ref{fig:curriculum}. Through this process, we obtain the final curriculum $\mathcal{C} = [l_1, ..., l_K]$, where $l_k$ is a natural language description of each subtask $\mathcal{M}_k$. Here, $l_k$ is grounded in the state variables of the environment to improve reward design and VLM evaluation. For example, in \texttt{Task 1: Approach Handles} (Fig.~\ref{fig:curriculum}), instead of a vague description such as “{\small \textit{each arm should get closer to its handle}},” the LLM-generated subtask explicitly uses the state variable \texttt{Euclidean distance to its respective handle}, and produces the description “{\small \textit{each arm minimizes its Euclidean distance to its target handle}}.”

\subsection{Reward Function Generation Module} \label{subsec: reward_generation}
For each subtask $\mathcal{M}_k$, CRAFT translates the natural language description $l_k$ into a reward function as executable code. Since the initial reward may not train a successful policy, CRAFT \textit{iteratively} refines the reward for the subtask. This subsection describes the generation of the initial reward $R_k^0$ from $l_k$. The superscript denotes the refinement index, with $R_k^j$ representing the reward after $j$ rounds of refinement. The refinement procedure is described in Section~\ref{subsec: reward_refinement}.

\begin{figure*}[!t]
    \centering
    \includegraphics[width=1.0\linewidth]{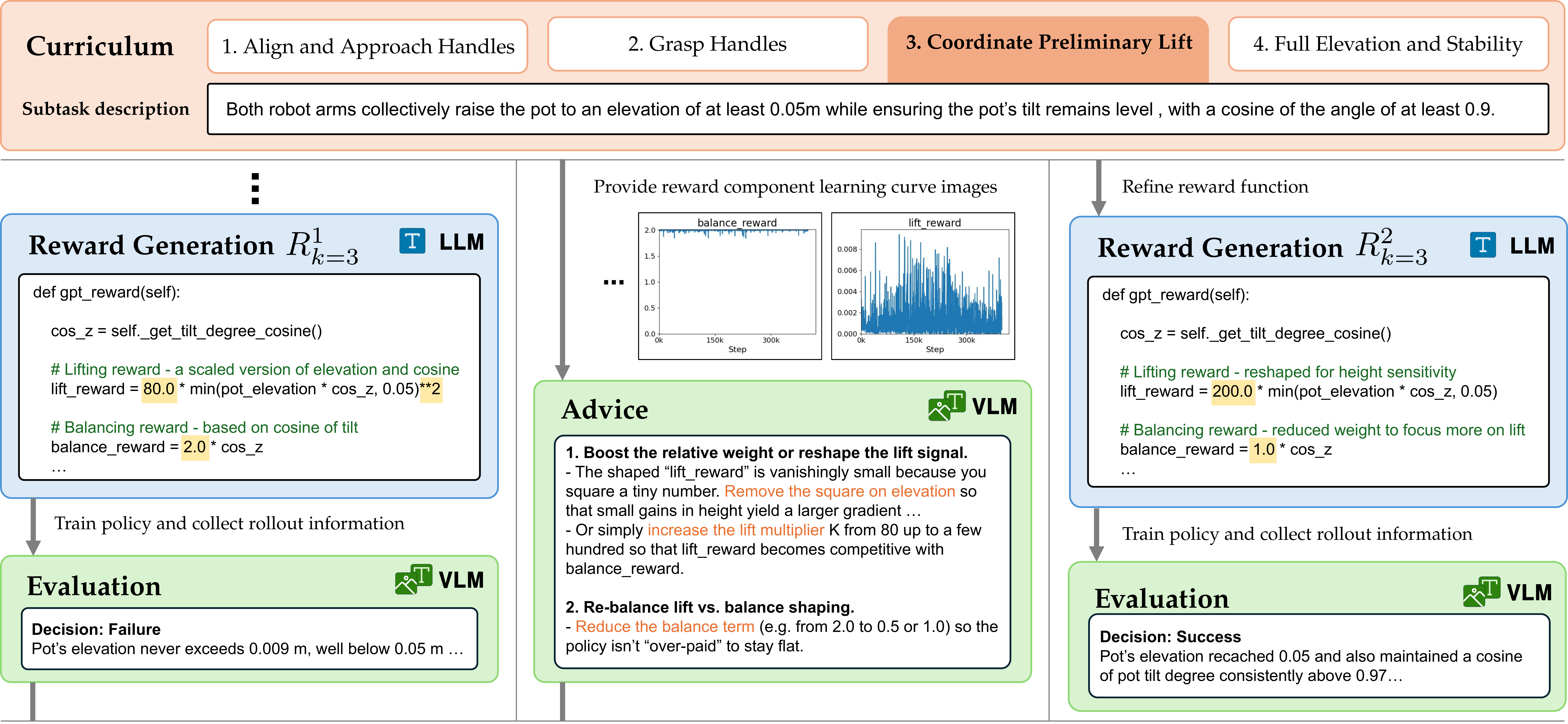}
    \caption{Example of reward refinement of subtask \texttt{Coordinate Preliminary Lift}. Through the first reward-refinement loop, $R^1_{k=3}$ was produced and the evaluation VLM marked the policy as a failure since the pot never reached the required elevation of \texttt{0.05 m}. The reward component learning curves were then passed to the advice VLM, which identified that \texttt{lift\_reward} was too weak compared to \texttt{balance\_reward}. It recommended removing the square on elevation, increasing the lift weight, and decreasing the balance weight. The revised reward $R^2_{k=3}$ reflects these changes: the square on elevation was removed, the lift weight increased from 80 to 200, and the balance weight decreased from 2 to 1. With this reward, the policy successfully achieved the \texttt{0.05 m} elevation and satisfied the subtask.}
    \label{fig:reward_refinement}
\end{figure*}

To generate $R_k^0$, we prompt the LLM with the environment description, the current subtask description $l_k$, and an example reward function with helper functions, which provide task-relevant structure and simplify processing of high-dimensional robot states. For instance, in the bimanual manipulation task shown in Fig.~\ref{fig:reward_refinement}, a helper function \texttt{self.\_get\_tilt\_degree\_cosine()} is provided to compute the cosine value of the pot’s tilt angle. We also include reward functions from earlier subtasks as the prompt context to exploit knowledge acquired in previous curriculum stages. In addition, we instruct the LLM to decompose the reward into interpretable components (e.g. \texttt{lift\_ reward}, \texttt{balance\_reward} in Fig.~\ref{fig:reward_refinement}), whose learning curves are later used for reward refinement in Section~\ref{subsec: reward_refinement}. Finally, we note that we design a reward for \textit{the team} rather than for specific agents, and train an initial policy $\pi_k^0$ using a CTDE MARL algorithm such as MAPPO~\cite{mappo}.

% To generate an initial reward function $R^0_k$, we prompt the LLM with the language description of the environment, current subtask description $l_k$, and an example reward function with helper functions, which provide task-relevant structure and simplify processing of high-dimensional robot states. For instance, in the bimanual manipulation task shown in Fig.~\ref{fig:reward_refinement}, a helper function \texttt{self.\_get\_tilt\_degree\_cosine()} is provided to compute the cosine value of the pot’s tilt angle. Since curriculum learning should exploit knowledge acquired in previous subtasks, we also include the reward functions from earlier subtasks as additional context in the prompt. In addition, when writing a reward function, we prompt the LLM to separate the reward into reward components (e.g. \texttt{lift\_ reward}, \texttt{balance\_reward} in Fig.~\ref{fig:reward_refinement}). These logs will later be provided as reward component learning curves for reward refinements in Section~\ref{subsec: reward_refinement}. After reward generation, we train a policy $\pi_k^0$ with the initial reward function $R_k^0$ via MARL, such as MAPPO~\cite{mappo}. We note that as we assume using CTDE MARL framework, we are designing a reward for a team, rather than specific agents.

\subsection{Policy Evaluation Module} \label{subsec: policy_eval}

% While one-shot LLM-generated rewards have shown success in single-agent domains \cite{eureka,curricullm}, we find them less reliable in multi-agent settings, where translating natural language task descriptions into effective reward functions often fails to induce the desired behavior. Therefore, we include a verification step to evaluate the trained policy $\pi^0_k$ and determine whether to advance to the next subtask or refine the current one. 

% When evaluating a robot policy, humans rely on visual trajectory rollouts to analyze the collaboration between robots and the progress in the subtask. Motivated by this, we use VLMs for evaluating the success or failure of $\pi_k^0$ by providing a sequence of snapshot images. Moreover, we also provide state variable trajectory data to make it easier for VLMs to conclude the subtask success. This design choice allows our method to be easily applied to various multi-robot platforms and arbitrary LLM-designed subtasks.

While one-shot LLM-generated rewards have shown success in single-agent domains~\cite{curricullm}, they can be less reliable in multi-agent settings, where reward design must capture both individual agent behaviors and the coordination required for team success. Therefore, we evaluate the trained policy $\pi^0_k$ before advancing to the next subtask and refine the current reward if the trained policy does not successfully complete subtask $M_k$. Motivated by how humans use visual trajectories to assess robot policies, we use a VLM to evaluate $\pi_k^0$ from a sequence of trajectory snapshots. We additionally provide state-variable trajectories for more reliable subtask success evaluation.

If the VLM judges the policy to be successful, we proceed to the next subtask $\mathcal{M}_{k+1}$. Otherwise, the VLM provides failure reasons, which are used to refine the current reward function $R_k^0$ toward the desired behavior. This image-and-state-based policy evaluation allows our method to be easily applied to diverse multi-robot platforms and arbitrary LLM-designed subtasks.

% In multi-robot environments, we describe the state of the robots through coordinates of positions and rotations. However LLM/VLMs lack an inherent understanding of 3D motion: while they perform reasonably on simple 2D tasks, they struggle with complex 3D environments. Instead of providing raw state variables, we extract task-relevant metrics (e.g., norms, or distances) that are easier for LLMs/VLMs to interpret. 
% % Moreover, collaboration between robots is difficult to capture through purely numerical state variables. 
% Moreover, when humans evaluate a policy, they rely on visual rollouts to analyze the collaboration between robots, and the progress in the subtask. Motivated by this, we include snapshot images of the rollout along with the state variables trajectory data. With this information, VLM is able to successfully evaluate the policy.

\subsection{Reward Refinement Module} \label{subsec: reward_refinement}

When the generated reward $R_k^0$ fails to induce the desired behavior, we should design a new reward function and retry the training process. A naive resampling with the same prompt~\cite{curricullm}, or just appending the previously attempted reward function~\cite{bosio2025synthesizing} requires extensive trial and error and are not scalable to high-dimensional multi-agent systems. 
% To overcome this, prior works suggested that a team of specialized models can be more effective than a single model for complex tasks~\cite{guo2024large,talebirad2023multi}. Following this,
To make refinement more targeted, we introduce two models: an \textit{advice} VLM and a \textit{refine} LLM, which enable more meaningful refinement.

% First, we prompt the advice VLM to generate free-form advice on how to refine the current reward function based on the current subtask description $l_k$, the previously attempted reward function $R_k^j$, and the failure reason obtained from policy evaluation module in Section~\ref{subsec: policy_eval}. In addition, we provide the learning curves of individual reward components, which indicate how each component evolved during training. Since these values often oscillate and are difficult to represent as long numerical sequences in text, we instead convert them into graphical images, offering the VLM a compact and interpretable representation. 

First, we use an advice VLM to generate free-form advice on how to refine the current reward function. We provide the advice VLM with the current subtask description $l_k$, the failed reward $R_k^j$, the failure reason from the policy-evaluation module in Section~\ref{subsec: policy_eval}, and the reward-component learning curves. These curves indicate how each reward component (e.g. \texttt{lift\_reward} in Fig.~\ref{fig:reward_refinement}) evolves during training. Since these values often oscillate and are difficult to summarize in text, we render them as plots and provide the VLM with a compact representation of the training dynamics.

% Second, in the refine LLM, we provide the previously attempted reward function $R_k^j$ and advice generated by the VLM. The refine LLM then produces a new refined reward $R_k^{j+1}$. Once a new refined reward $R^{j+1}_k$ is generated, we train a corresponding policy $\pi^{j+1}_k$ and iterate through the VLM-guided reward–refinement loop. We allow a maximum of $J$ refinement iterations. If the policy still fails after $J$ iterations, the LLM selects the best policy among $\pi_k^0, ... , \pi_k^J$ using rollout information and proceeds to the next subtask. In our experiments, we found that $J=3$ iterations were sufficient to generate a successful reward. As one iteration consist of 3 LLM/VLM queries for reward generation, evaluation, and advice, we can learn one subtask under 9 LLM/VLM queries, thereby minimizing the computational overhead associated with LLM/VLM inference. An example of this refinement process is shown in Fig.~\ref{fig:reward_refinement}.

Second, the refine LLM takes the previously attempted reward function $R_k^j$ and the advice generated by the VLM. The refine LLM then produces a new refined reward $R_k^{j+1}$. We train a corresponding policy $\pi_k^{j+1}$ with this reward and iterate over the VLM-guided reward-refinement loop for at most $J$ iterations. If all attempts fail, the LLM selects the best policy among $\pi_k^0,\ldots,\pi_k^J$ using rollout information, and the curriculum proceeds to the next subtask. In our experiments, $J=3$ was sufficient to obtain successful rewards, requiring at most 9 LLM/VLM queries per subtask after initial reward generation. An example of this refinement process is shown in Fig.~\ref{fig:reward_refinement}.

Once we obtain the best policy $\pi_k^*$ for subtask $\mathcal{M}_k$, we proceed to the next subtask $\mathcal{M}_{k+1}$. To train the policy $\pi_{k+1}$ for $\mathcal{M}_{k+1}$, we initialize its weights using $\pi_k^*$. This warm-starting strategy (Appendix~\ref{subsec: craft_detail}) helps retain behaviors from earlier subtasks while learning more complex skills. 

\section{Experiments}

In this section, we evaluate our framework in several collaborative multi-robot scenarios. We consider a bimanual manipulation task and multi-quadruped navigation tasks that require complex long-horizon coordination. We show that CRAFT can learn coordination that is challenging to achieve with vanilla MARL even with human-designed or LLM-generated reward functions. We also validate our quadruped navigation task in hardware to show our policy transfers to real-world.

\subsection{Experiment Setup}

\subsubsection{Simulation Environments}

\begin{figure}
    \centering
    \begin{subfigure}[b]{0.31\linewidth}
        \centering
        \includegraphics[width=\linewidth,height=0.12\textheight]{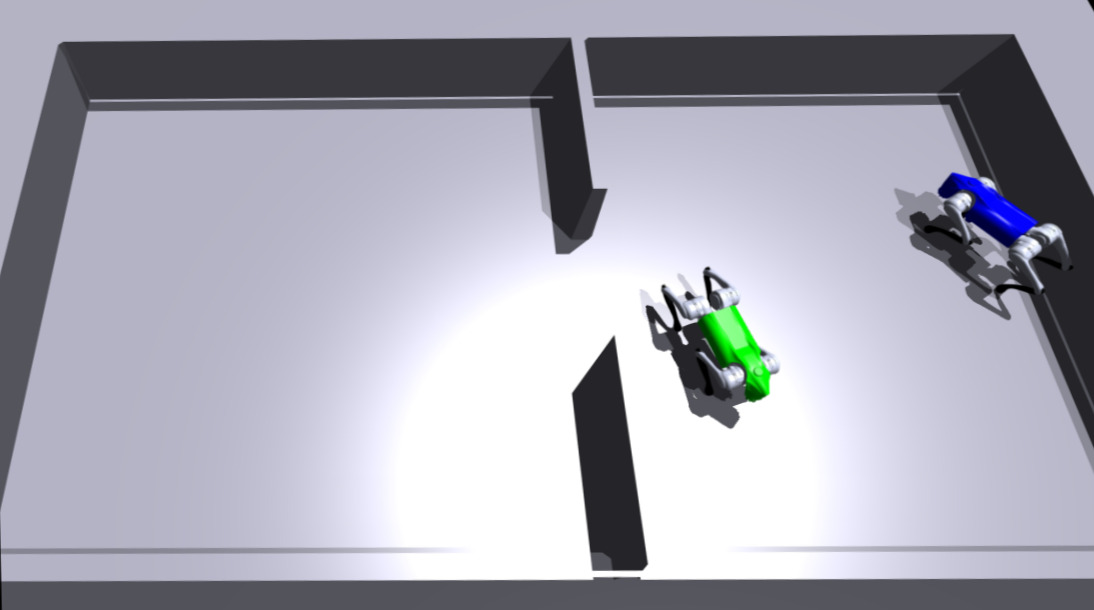}
    \end{subfigure}
    \hfill
    \begin{subfigure}[b]{0.31\linewidth}
        \centering
        \includegraphics[width=\linewidth,height=0.12\textheight]{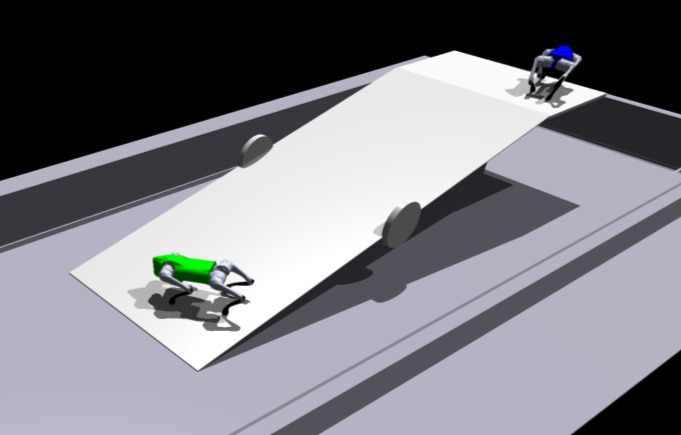}
    \end{subfigure}
    \hfill
    \begin{subfigure}[b]{0.31\linewidth}
        \centering
        \includegraphics[width=\linewidth,height=0.12\textheight]{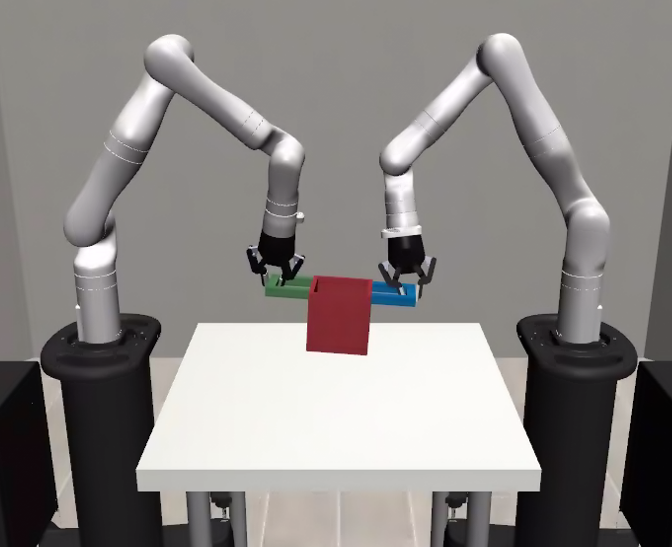}
    \end{subfigure}
    \caption{Illustrative snapshot showing successful execution of multi-agent coordination tasks by policies trained by CRAFT.}
    \label{fig:env_figure}
\end{figure}

% \begin{wrapfigure}{r}{0.5\textwidth}
% \vspace{-50pt}
%     \centering
%     \scriptsize
%     \begin{subfigure}[b]{0.31\linewidth}
%         \centering
%         \includegraphics[width=\linewidth,height=0.07\textheight]{corl_2026_template_submission/figures/go2gate_env.jpeg}
%     \end{subfigure}
%     \hfill
%     \begin{subfigure}[b]{0.31\linewidth}
%         \centering
%         \includegraphics[width=\linewidth,height=0.07\textheight]{corl_2026_template_submission/figures/go2seesaw_env.jpeg}
%     \end{subfigure}
%     \hfill
%     \begin{subfigure}[b]{0.31\linewidth}
%         \centering
%         \includegraphics[width=\linewidth,height=0.07\textheight]{corl_2026_template_submission/figures/lift_example.png}
%     \end{subfigure}
%     \caption{Illustrative snapshot showing successful execution of multi-agent coordination tasks by policies trained by CRAFT.}
%     \label{fig:env_figure}
%     \vspace{-10pt}
% \end{wrapfigure}
 
We evaluate CRAFT in decentralized multi-agent coordination environments (Fig.~\ref{fig:env_figure}):

\begin{itemize}[leftmargin=*]
    \item \textbf{Quadruped Gate} (Gate): Two quadrupeds should coordinate to pass the narrow gate without collision~\cite{mqe}. We define a task as success when both agents have passed the gate. Each quadruped has a continuous action space of $x$- and $y$-axis velocities, and yaw velocity.
    \item \textbf{Quadruped Seesaw} (Seesaw): Two quadrupeds should coordinate on the seesaw so that one agent climbs up to the platform at the top while the other balances it~\cite{mqe}. We define a task as success when one of the agents reaches the target platform. The action space is identical to Gate environment.
    \item \textbf{Two Arm Lift} (Lift): Two robot arms should coordinate in lifting the pot together while keeping the pot level~\cite{robosuite2020}. We define a task as success when the pot is lifted above $0.1$m and no more tilted than 30 degrees. Each arm has a continuous action space, which consists of six end-effector delta pose, position and orientation changes relative to the current pose, and one gripper state. 
\end{itemize}
% \noindent \textbf{Quadruped Gate} (Gate): Two quadrupeds should coordinate to pass the narrow gate without collision~\cite{mqe}. We define a task as success when both agents have passed the gate. Each quadruped has a continuous action space of $x$- and $y$-axis velocities, and yaw velocity. \\
% \noindent \textbf{Quadruped Seesaw} (Seesaw): Two quadrupeds should coordinate on the seesaw so that one agent climbs up to the platform at the top while the other balances it~\cite{mqe}.
% %to utilize the seesaw for one agent climbing up to the platform while the other agent balances the seesaw~\cite{mqe}. 
% We define a task as success when one of the agents reaches the target platform. The action space is identical to Gate environment. \\
% \noindent \textbf{Two Arm Lift} (Lift): Two robot arms should coordinate in lifting the pot together while keeping the pot level~\cite{robosuite2020}. We define a task as success when the pot is lifted above 0.1m and no more tilted than 30 degrees. Each arm has a continuous action space, which consists of six end-effector delta pose, position and orientation changes relative to the current pose, and one gripper state. 

In these environments, each agent observes its own state and partial information about the other agent, but not the other agent's actions or intentions. For example, each quadruped observes the other quadruped's position and orientation, but does not know their future actions. This decentralized execution creates a difficult exploration problem, where successful outcomes require synchronized multi-agent actions and uncoordinated exploration often yields weak learning signals, especially without demonstrations or offline data. We use MAPPO~\cite{mappo}, which employs decentralized execution, to evaluate CRAFT's ability to learn such coordination.
% In these environments, each agent can observe its own state as well as partial information about the other agent. However, agents do not have access to the other agent’s actions or intentions. For example, in the quadruped environments, each agent can observe the positions and orientations of itself and the other agent, but does not know how the other agent will move from its current state. Thus, decentralized agents must learn to infer and adapt to one another's behavior from local observations, making coordination learning challenging.
% Such decentralized policies induce non-stationarity during MARL training, making coordination learning challenging. 
% Therefore, to our knowledge, no prior RL methods have succeeded in solving these tasks reliably without expert demonstrations.
% These environments are exceptionally challenging for MARL as they demand long-horizon  coordination and decentralized control under partial observability. To our knowledge, no prior RL methods without expert demonstrations have succeeded in solving them reliably.
More details about the environment and experiment settings can be found in Appendix~\ref{sec: exp_detail}.

\subsubsection{Baselines \& Ablation Study}

We compare CRAFT with the following baselines and ablations. \textbf{Env Reward} uses the environment-provided reward, while \textbf{Example Reward} uses the example reward provided to the reward generation module of CRAFT. Both baselines are trained without any staged curricula. \textbf{From Scratch} uses the final reward functions generated by CRAFT, but trains MARL without a curriculum and keep the reward function fixed during training to assess the impact of staged curriculum training.
We further compare CRAFT with prior LLM-based reward generation and refinement methods. 
% We further compare CRAFT with \textbf{Eureka}~\citep{eureka} and \textbf{CurricuLLM}~\citep{curricullm}. 
% \textbf{Eureka}~\citep{eureka} uses LLMs to iteratively refine rewards, but does not generate a curriculum. In each refinement iteration, Eureka samples multiple reward functions, trains a policy for each candidate, selects the best-performing reward using a task-specific scalar metric predefined by a human engineer, and refines from that reward in the next iteration. \textbf{CurricuLLM}~\citep{curricullm} uses LLM-generated curricula and rewards, but does not perform iterative reward refinement. Instead, it resamples new reward functions from the reward module, relying on the stochasticity of the LLM to generate effective reward functions. 
\textbf{Eureka}~\citep{eureka} uses LLMs to iteratively generate and refine rewards, but trains directly on the target task without a curriculum. \textbf{CurricuLLM}~\citep{curricullm} uses LLM-generated curricula and rewards, but does not perform iterative reward refinement. 
These baselines use the same environment descriptions, helper functions, and language model access where applicable. More details can be found in Appendix~\ref{subsec: baseline_detail}.
% These baselines allow us to evaluate the importance of curriculum learning and reward refinement in CRAFT.
% \textcolor{red}{Since LLM-based method can produce degenerate reward or curricula due to stochastic generation, we run each method 5 times and report the best 3 trials in the main results.}

Since LLM-based methods can vary across trials due to their stochastic generation nature, we run each method $5$ times and report the top-three trials for each method. To further analyze the reliability of generated curricula and rewards, we report two additional metrics: \textit{curriculum success ratio}, defined as the fraction of $5$ trials in which the final policy achieves a nonzero success rate, and \textit{maximum success rate}, defined as the maximum success rate achieved across trials.

% \begin{figure*}[t]
%     \centering
%     \begin{subfigure}[b]{0.32\linewidth}
%         \centering
%         \includegraphics[width=\linewidth]{figures/gate_total_success_rate.pdf}
%         \caption{Quadruped Gate}
%     \end{subfigure}
%     \hfill
%     \begin{subfigure}[b]{0.32\linewidth}
%         \centering
%         \includegraphics[width=\linewidth]{figures/seesaw_total_success_rate.pdf}
%         \caption{Quadruped Seesaw}
%     \end{subfigure}
%     \hfill
%     \begin{subfigure}[b]{0.32\linewidth}
%         \centering
%         \includegraphics[width=\linewidth]{figures/lift_total_success_rate.pdf}
%         \caption{Two Arm Lift}
%     \end{subfigure}
%     \caption{Success Rate of the top-three curricula from each environment. Each environment is evaluated by 100 random initial conditions. CRAFT achieves highest success rate on every environment, demonstrating its ability to learn complex coordination tasks that are challenging to learn without curriculum or well-crafted reward functions.}
%     \label{fig:total_success_rate}
%     % \vspace{-1.0em}
% \end{figure*}

\newcommand{\tallrow}{\rule{0pt}{3.8ex}}
\begin{figure*}[t]
    \centering
    \begin{minipage}{0.60\textwidth}
        \centering
        \scriptsize
        \begin{tabular}{l|c|c|c}
            \toprule
            & \makecell{Gate} 
            & \makecell{Seesaw} 
            & \makecell{Lift} \\
            \midrule
            Env Reward & $0.60 \pm 1.20$ & $0.00 \pm 0.00$ & $1.33 \pm 1.89$ \\
            Example Reward & $0.00 \pm 0.00$ & $4.19 \pm 5.08$ & $9.33 \pm 3.77$\\
            From Scratch & $15.00 \pm 21.21$ & $24.67 \pm 6.60$ & $8.0 \pm 11.31$\\
            Eureka~\citep{eureka} & $\mathbf{83.33 \pm 3.57}$ & $15.00 \pm 11.34$ & $41.33 \pm 12.36$\\
            CurricuLLM~\citep{curricullm} & $27.75 \pm 35.59$ & $21.82 \pm 24.13$ & $25.33 \pm 35.83$\\
            CRAFT (Ours) & $\mathbf{78.33 \pm 9.88}$ & $\mathbf{59.33 \pm 10.62}$ & $\mathbf{94.67 \pm 4.99}$\\
            \bottomrule
        \end{tabular}
        \captionof{table}{Success Rate(\%) of the top-three out of $5$ trials from each environment. Each environment is evaluated by 100 random initial conditions. CRAFT achieves the highest success rates in Seesaw and Lift, and remains competitive with the strongest baseline in Gate.}
        \label{tab:success_rate}
    \end{minipage}
    \hfill
    \begin{minipage}{0.37\textwidth}
        \centering
        \scriptsize
        \begin{subfigure}[b]{0.49\linewidth}
            \centering
            \includegraphics[width=\linewidth]{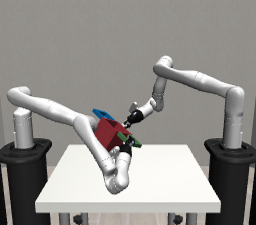}
            \caption{Eureka} % ($52\%$)
        \end{subfigure}
        \hfill
        \begin{subfigure}[b]{0.49\linewidth}
            \centering
            \includegraphics[width=\linewidth]{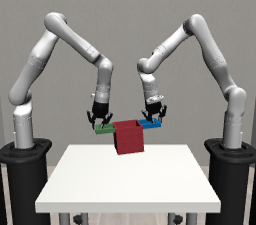}
            \caption{CRAFT} % ($100\%$)
        \end{subfigure}
        \captionof{figure}{Highest success rate Lift policies from Eureka and CRAFT. Eureka shows unnatural joint configurations, while CRAFT demonstrates more physically feasible motions.}
        \label{fig:comparison_twoarmlift}
    \end{minipage}
\end{figure*}

\subsection{Simulation Results} \label{subsec: results}
\textbf{Q1. Can our coach train MARL policies that require complex coordination between multiple robots?} 
% Across all three environments, CRAFT learns \emph{successful policies without expert demonstrations or hand-crafted curricula}. 
% Across all three environments, CRAFT learns \emph{successful policies and outperforms existing LLM-based reward design and curriculum design methods}. 
Across all three environments, CRAFT learns successful policies without expert demonstrations or hand-crafted curricula.
In Table~\ref{tab:success_rate}, CRAFT achieves the highest success rates in Seesaw and Lift, while remaining competitive with the strongest baseline in Gate. In contrast, Env Reward achieves near-zero success in most experiments despite using the dense rewards designed by the environment developers. Similarly, the Example Reward provided to CRAFT achieves less than $10\%$ success in all environments which indicates that the example reward alone is insufficient. Nevertheless, CRAFT builds successful curricula and refines rewards to train policies with substantially higher success rates.
Table~\ref{tab:SR_curriculum} shows that the best policies obtained from CRAFT can reach maximum success rates of up to $92\%$ in Gate, $72\%$ in Seesaw, and $100\%$ in Lift. Together, these results demonstrate that CRAFT can learn effective coordination policies across diverse multi-robot tasks.
\textbf{Q2. Is curriculum generation necessary for learning effective coordination policies?}
Compared to From Scratch, as shown in Table~\ref{tab:success_rate}, CRAFT achieves higher success rates across all environments, indicating that CRAFT-optimized rewards alone are insufficient without curriculum. Compared to Eureka, CRAFT achieves larger gains in Seesaw and Lift, which suggests that reward refinement alone becomes less sufficient for multi-stage coordination tasks. We further observe a qualitative limitation of Eureka in Lift. Directly optimizing for the final task without building the intermediate skills can lead to undesirable behaviors as shown in Fig.~\ref{fig:comparison_twoarmlift} and Appendix~\ref{subsec: additional_analysis}. Eureka lifts the pot, but does so with unnatural joint configurations that are difficult to transfer to real hardware. This suggests that reward search on the final task can overemphasize terminal success, such as elevating the pot, while neglecting intermediate behaviors needed for physically plausible execution, such as aligning with the handles and establishing stable grasps. In contrast, CRAFT's staged curriculum builds these skills progressively, and guides policies toward more realistic coordination.

% In addition, while no\_curriculum achieves a comparable success rate to CRAFT in the Quadruped Seesaw environment, we note that it benefits from the well-designed reward function produced by CRAFT. In contrast, env\_reward and example\_reward fail to learn effective coordination behaviors, resulting in much lower success rates. These results indicate that CRAFT’s reward-refinement loop contributes to generating well-designed rewards that go beyond the capabilities of any provided reward functions.
% Moreover, we note that high policy-SR of no\_curriculum policies in Quadruped environments arises from the well-designed reward functions produced by CRAFT’s reward refinements.

% \begin{figure}[t]
%     \centering
%     \begin{subfigure}[b]{0.31\linewidth}
%         \centering
%         \includegraphics[width=\linewidth,height=0.12\textheight]{figures/go2gate_partial.png}
%     \end{subfigure}
%     \hfill
%     \begin{subfigure}[b]{0.31\linewidth}
%         \centering
%         \includegraphics[width=\linewidth,height=0.12\textheight]{figures/go2seesaw_partial.png}
%     \end{subfigure}
%     \hfill
%     \begin{subfigure}[b]{0.31\linewidth}
%         \centering
%         \includegraphics[width=\linewidth,height=0.12\textheight]{figures/lift_partial_success.png}
%     \end{subfigure}
%     \caption{Illustrative snapshot of policies trained with env\_reward without curriculum. The policy shows suboptimal behaviors, such as only one agent passing the gate, agent failing to balance the seesaw, or only managed to grasp the pot, rather than lifting it.}
%     \label{fig:suboptimal}
% \end{figure}

\begin{figure*}[t]
    \centering

    \begin{minipage}{0.60\textwidth}
        \centering
        \scriptsize
        \setlength{\tabcolsep}{4pt}
        \begin{tabular}{c|c|c|c}
            \toprule
             & & CRAFT & CurricuLLM~\citep{curricullm}  \\
             \midrule
            \multirow{2}{*}{\makecell{\textbf{\textbf{Gate}}}} 
            & Curriculum Success Ratio & $\mathbf{4/5}$ & $2/5$ \\
            & Maximum Success Rate & $\mathbf{92\%}$ & $78\%$ \\
             \midrule
            \multirow{2}{*}{\makecell{\textbf{Seesaw}}}
            & Curriculum Success Ratio & $\mathbf{3/5}$ & $2/5$ \\
            & Maximum Success Rate & $\mathbf{72\%}$ & $55\%$ \\
             \midrule
            \multirow{2}{*}{\makecell{\textbf{Lift}}}
            & Curriculum Success Ratio & $\mathbf{4/5}$ & $1/5$ \\
            & Maximum Success Rate & $\mathbf{100\%}$ & $76\%$ \\
            \bottomrule
        \end{tabular}
        \captionof{table}{Curriculum success ratio measures the fraction of curricula with nonzero success across $5$ trials, while maximum success rate reports the best policy over $5$ trials. CRAFT produces more reliable curricula than CurricuLLM, and its higher maximum success rate shows the benefit of reward refinement for generating better policies.}
        \label{tab:SR_curriculum}
    \end{minipage}
    \hfill
    \begin{minipage}{0.38\textwidth}
        \centering
        \scriptsize
        \begin{tabular}{c c c c}
            \toprule
            \multicolumn{2}{c}{} & \multicolumn{2}{c}{VLM Decision} \\
            \multicolumn{2}{c}{} & Positive & Negative \\
            \midrule
            \multirow{2}{*}{\rotatebox{90}{Actual}}
            & Positive \tallrow
            & \multicolumn{1}{>{\columncolor[rgb]{0.90,1.00,0.90}}c}{$7 / 35$ (TP)}
            & \multicolumn{1}{>{\columncolor[rgb]{1.00,0.90,0.90}}c}{$5/35$ (FN)} \\
            & Negative \tallrow
            & \multicolumn{1}{>{\columncolor[rgb]{1.00,0.90,0.90}}c}{$0/35$ (FP)}
            & \multicolumn{1}{>{\columncolor[rgb]{0.90,1.00,0.90}}c}{$23/35$ (TN)} \\
            \bottomrule
        \end{tabular}
        \captionof{table}{Confusion matrix for VLM policy evaluation on final task rollouts, aggregated across environments. 
        % Positive and negative denote successful and unsuccessful policy rollouts, respectively. 
        TP, FN, FP, and TN indicate true positives, false negatives, false positives, and true negatives. 
        The VLM achieves $85.71\%$ accuracy, with all errors corresponding to false negatives.
        }
        \label{tab:confusion_matrix_rates}
    \end{minipage}
\end{figure*}

% \textbf{\ref{Q: refinement}: The VLM-guided reward-refinement loop in CRAFT improves curricula reliability.} 
\textbf{Q3. How does the reward refinement module help learning successful policies?} 
% To evaluate the consistency of generated curricula and reward functions, we report the \textit{effective curricula ratio}, defined as the fraction of trials in which the final policy achieves a nonzero success rate, and the \textit{maximum success rate}, defined as the best success rate achieved among multiple trials in Table~\ref{tab:SR_curriculum}. 
As shown in Table~\ref{tab:SR_curriculum}, CRAFT achieves both higher curriculum success ratio and maximum success rate than CurricuLLM. The higher curriculum success ratio indicates that VLM-guided reward refinement contributes to reliable subtask training, which comes from more consistent generation of subtask rewards. The higher maximum success rate further shows that reward refinement not only improves reliability, but also helps produce better-performing policies. Although CurricuLLM can occasionally learn a successful policy, its lower curriculum success ratio indicates less reliable curriculum generation, and degrades its average performance as shown in Table~\ref{tab:success_rate}. 

\textbf{Q4. How reliable is the VLM decision in the policy evaluation module?} We investigate whether the VLM-based policy evaluation introduced in Section~\ref{subsec: policy_eval} suffers from hallucinations. Specifically, we compare the VLM's judgments on final-task rollouts against the ground-truth success in Table~\ref{tab:confusion_matrix_rates}. 
% We focus on the final task because the intermediate subtasks vary across curricula, making it difficult to define a consistent success criterion for each subtask. 
This provides a proxy for evaluation reliability, since intermediate subtasks vary across generated curricula and are therefore harder to benchmark consistently.
The VLM makes the correct decision in $85.71\%$ of the cases, while the remaining $14.29\%$ are all false negatives. This indicates that the VLM behaves conservatively by occasionally rejecting successful policies. Such errors are tolerable in CRAFT since a false negative simply results in continuing the training with a refined reward. Even if later refinement iterations fail to produce better policies, CRAFT retains previously evaluated successful policies and can select the best one across the $J$ refinement iterations.

\subsection{Hardware Validation}
We validate our method by transferring the learned policy to real robots on the Quadruped Gate task. The policy is zero-shot transferred from training in simulation with CRAFT to hardware. We deploy the policy on one Unitree Go2 and one Unitree Go1 robot. While each robot receives both agents' positions and orientations from a motion capture system, policy execution is fully decentralized.
% Our decentralized navigation policy receives the positions and orientations of both quadrupeds, as well as the position of the gate. 
% The robots’ positions and orientations are measured using a motion capture system and provided as inputs to the policy. 
As shown in Fig.~\ref{fig:overall_method}, our zero-shot transferred policy is successfully deployed in real world and achieves a success rate of 65\% over 20 runs, serving as a solid baseline given that no finetuning is performed after simulation training. More details can be found in Appendix~\ref{subsec: hardware_exp}.

\section{Conclusions and Limitations}

In this paper, we proposed \textbf{CRAFT}, a framework to train long-horizon multi-robot coordination tasks. We leveraged the reasoning capabilities of foundation models, with LLMs and VLMs acting as a ``coach'' to decompose complex behaviors into curricula of subtasks, each paired with an LLM-generated reward function. A VLM-guided reward-refinement loop further improves these rewards and produces semantically rich reward signals that enable successful learning in long-horizon coordination tasks. Through our experiments in multi-quadruped navigation and bimanual manipulation, we demonstrated that our method is effective and also transferable to real hardware.

A key limitation of CRAFT is the variability across trials as the stochasticity of foundation models can occasionally produce failures. However, this is manageable in simulation as we can make multiple training attempts and deploy only successful policies in real-world. This stochasticity can also be beneficial, as diverse curricula and advices may lead to more creative solutions. In addition, we found that the VLM-based policy evaluation module relies more heavily on numerical state trajectories than visual rollout information. This may limit the VLM's ability to detect visually grounded failures such as poor spatial coordination. Future work can address this by more structured visual prompts, temporally annotated rollout images, or video-based VLMs that reason directly over motion. 

% %===============================================================================

% \section{Citations}
% \label{sec:citations}

% 	Citations can be made using either \textbackslash citep\{\} or \textbackslash citet\{\}, depending from the appropriateness. To avoid the citation moving to the next line, it is often a good practice to replace the space before with a tilde (\~{}) character.
% 	Example 1: ``CoRL is the best conference ever~\citep{Gauss1857}.''
% 	Example 2: ``\citet{Lagrange1788} proved, both theoretically and numerically, that CoRL is the best conference ever.''

% %===============================================================================

\clearpage
% The acknowledgments are automatically included only in the final and preprint versions of the paper.
\acknowledgments{
The authors are grateful to Yuman Gao, Prof. Koushil Sreenath and the Hybrid Robotics Lab for their support on the hardware experiments.
% If a paper is accepted, the final camera-ready version will (and probably should) include acknowledgments. All acknowledgments go at the end of the paper, including thanks to reviewers who gave useful comments, to colleagues who contributed to the ideas, and to funding agencies and corporate sponsors that provided financial support.
}

%===============================================================================

\bibliography{example}  % .bib

\clearpage
\appendix
\input{appendix}

\end{document}

%% file: appendix.tex
\section{Details on Experiments} \label{sec: exp_detail}

\subsection{Implementation Details of CRAFT} \label{subsec: craft_detail}
We use MAPPO~\cite{mappo}, which employs decentralized execution with only local observations, to evaluate CRAFT's ability to learn decentralized coordination. MAPPO is implemented with OpenRL~\cite{openrl} for the multi-quadruped navigation, and skrl~\cite{serrano2023skrl} for the bimanual manipulation task. We utilize \texttt{gpt-4o-2024-08-06} as our LLM models and \texttt{o4-mini-2025-04-16} as our VLM models. 

When moving from one subtask to the next, we initialize the policy $\pi_{k+1}$ for subtask $\mathcal{M}_{k+1}$ using the weights of policy $\pi_k$ from subtask $\mathcal{M}_k$, then train it with the reward function $R_{k+1}$. However, because the reward generation LLM receives the history of rewards as context, $R_{k+1}$ may share components with $R_k$. This can cause a loss of plasticity~\cite{abbas2023loss,dohare2024loss} leading the policy $\pi_{k+1}$ to converge to a local optimum close to policy $\pi_k$, which can limit its adaptation to the new subtask.

To mitigate this issue, we apply two resets at the beginning of each subtask. First, we reset the exploration parameters of the policy network, which correspond to the policy standard deviation layer in our MAPPO implementation, back to their initial values. This allows the policy to explore alternative actions while preserving previously learned behavior, and help it escape local optima induced by the earlier reward function. Second, we reset the value network with random weights~\cite{cable-towed-qudruped}. By reinitializing only the value network while preserving the actor weights, the actor can exploit the previous policy, while the critic learns value estimates under the new reward function.

\subsection{Implementation Details of Baselines} \label{subsec: baseline_detail}
For all baseline experiments, we use the same MAPPO implementation and language model, when applicable, as in CRAFT, as described in Appendix~\ref{subsec: craft_detail}. To make the comparison as fair as possible, we provide LLM-based baselines with the same environment descriptions, state variables, helper functions, and example reward information used by CRAFT whenever the baseline method supports these inputs. 
For CurricuLLM, we use the same curriculum-generation and reward-generation prompt structure as CRAFT, including the same environment and state-variable descriptions, but remove the VLM-guided reward-refinement loop.
For Eureka, the original method provides the full Python source code of the environment when generating rewards. In practice, however, the full environment code can exceed the language model's context window. Eureka therefore uses an automatic script to extract useful environment code snippets~\citep{eureka}. Instead of implementing this automatic extraction procedure, we provide Eureka with the helper functions and environment descriptions used in CRAFT as the environment code snippets. Eureka originally performs evolutionary reward search with $5$ refinement iterations and $16$ reward samples per iteration. To match the training budget of CRAFT, we instead use $3$ refinement iterations with $3$ samples per iteration, resulting in a total of $9$ policy-training calls. For Eureka, which requires a scalar metric to select reward candidates at each iteration, we use the task success rate as the selection metric.

\subsection{Simulation Environment}
This section describes the observation and action spaces for each environment. Since all environments operate under a decentralized setting, the policy’s action space is restricted to agent-wise actions.

\textbf{Quadruped Gate}
\vspace{-0.5em}
\begin{itemize}[leftmargin=*]
    \item Observation space: agent ID; ego agent pose (position and orientation); other agent pose; and gate position.
    \vspace{-3pt}
    \item Action space: ego agent linear velocities in the $x$ and $y$ directions and angular velocity around the yaw axis.
\end{itemize}

\textbf{Quadruped Seesaw}
\vspace{-0.5em}
\begin{itemize}[leftmargin=*]
    \item Observation space: agent ID; ego agent pose; and other agent pose.
    \vspace{-3pt}
    \item Action space: ego agent linear velocities in the $x$ and $y$ directions and angular velocity around the yaw axis.
\end{itemize}

\textbf{Two-Arm Lift}
\vspace{-0.5em}
\begin{itemize}[leftmargin=*]
    \item Observation space: ego agent end-effector (EE) pose; ego agent gripper state; ego agent object handle pose; relative position of the object handle with respect to the ego agent EE; other agent’s EE pose; relative position of the object handle with respect to the other agent’s EE; and gripper state. (All poses in the observation are given in ego agent's base frame, not in the world frame. The ego agent will only know other's information with respect to itself.)
    \vspace{-3pt}
    \item Action space: delta pose of the end-effector and gripper state.
    \vspace{-3pt}
    \item The pose of the pot is randomized at every start of the episode.
\end{itemize}

\subsection{Additional Analysis on Experiments} \label{subsec: additional_analysis}

% \begin{table}[]
%     \centering
%     \begin{tabular}{l|c|c|c}
%             \toprule
%             & \makecell{Gate} 
%             & \makecell{Seesaw} 
%             & \makecell{Lift} \\
%             \midrule
%             Env Reward & $0.60 \pm 1.20$ & $0.00 \pm 0.00$ & $1.33 \pm 1.89$ \\
%             Example Reward & $0.00 \pm 0.00$ & $4.19 \pm 5.08$ & $9.33 \pm 3.77$\\
%             From Scratch & $15.00 \pm 21.21$ & $24.67 \pm 6.60$ & $8.0 \pm 11.31$\\
%             Eureka~\citep{eureka} & $\mathbf{83.33 \pm 3.57}$ & $15.00 \pm 11.34$ & $41.33 \pm 12.36$\\
%             CurricuLLM~\citep{curricullm} & $27.75 \pm 35.59$ & $21.82 \pm 24.13$ & $25.33 \pm 35.83$\\
%             CRAFT (Ours) & $\mathbf{78.33 \pm 9.88}$ & $\mathbf{59.33 \pm 10.62}$ & $\mathbf{94.67 \pm 4.99}$\\
%             \bottomrule
%     \end{tabular}
%     \caption{Success Rate(\%) of the top-three out of $5$ trials from each environment. Each environment is evaluated by 100 random initial conditions. CRAFT achieves the highest success rates in Seesaw and Lift, and remains competitive with the strongest baseline in Gate.}
%     \label{tab:placeholder}
% \end{table}

\begin{figure}[h]
    \centering
    \begin{subfigure}[b]{0.31\linewidth}
        \centering
        \includegraphics[width=\linewidth,height=0.12\textheight]{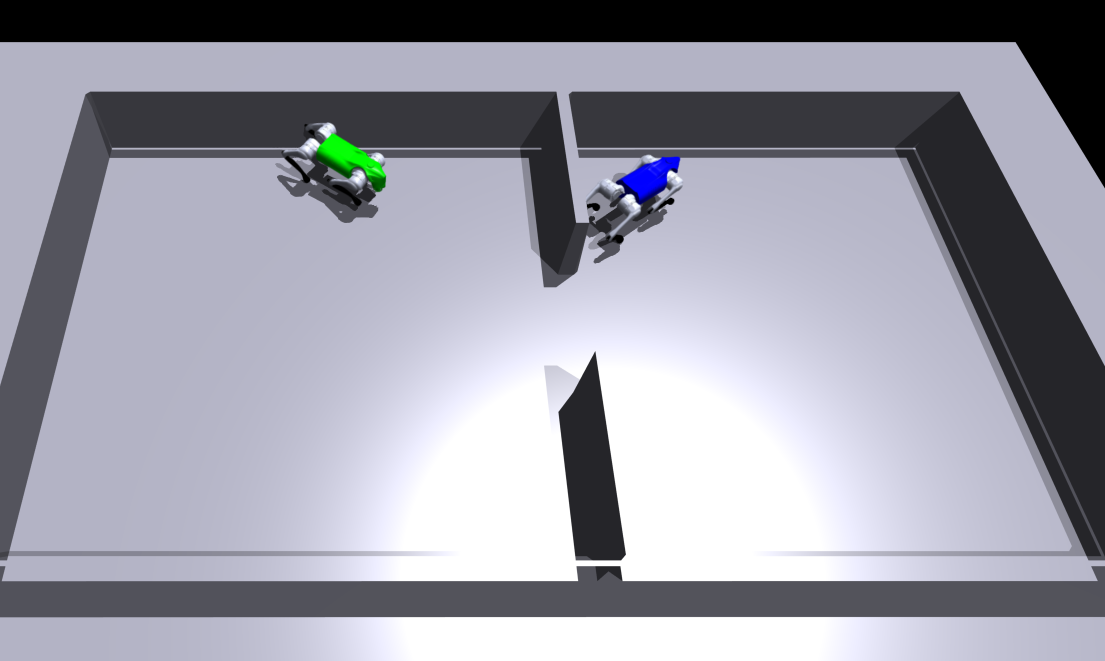}
    \end{subfigure}
    \hfill
    \begin{subfigure}[b]{0.31\linewidth}
        \centering
        \includegraphics[width=\linewidth,height=0.12\textheight]{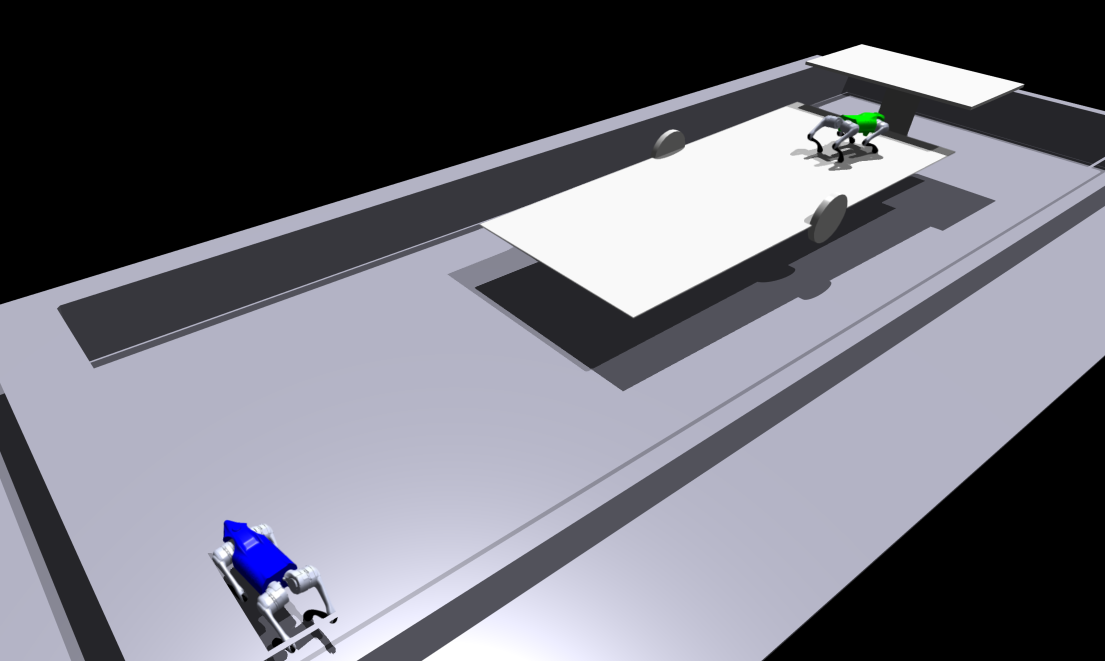}
    \end{subfigure}
    \hfill
    \begin{subfigure}[b]{0.31\linewidth}
        \centering
        \includegraphics[width=\linewidth,height=0.12\textheight]{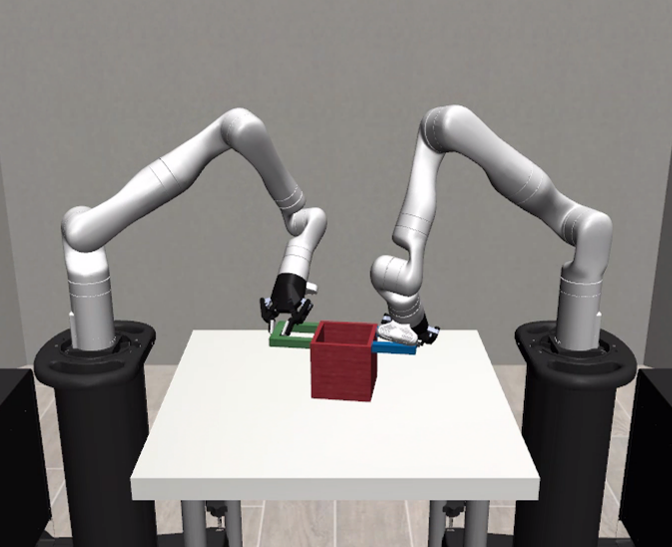}
    \end{subfigure}
    \caption{Illustrative snapshot of policies trained with Env Reward without curriculum. The policy shows suboptimal behaviors, such as only one agent passing the gate, agent failing to balance the seesaw, or only managing to grasp the pot, rather than lifting it.}
    \label{fig:suboptimal_env_reward}
\end{figure}

\begin{figure}[h]
    \captionsetup[subfigure]{labelformat=empty, font=small, skip=1pt}
    \centering
    \begin{subfigure}[b]{0.19\linewidth}
        \centering
        \includegraphics[width=\linewidth]{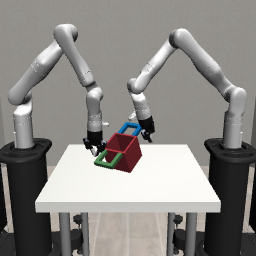}
        \caption{SR 8\% (F)}
    \end{subfigure}
    \hfill
    \begin{subfigure}[b]{0.19\linewidth}
        \centering
        \includegraphics[width=\linewidth]{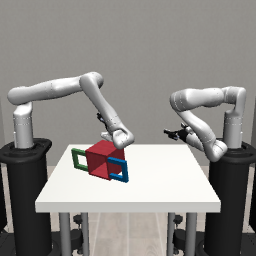}
        \caption{SR 12\% (F)}
    \end{subfigure}
    \hfill
    \begin{subfigure}[b]{0.19\linewidth}
        \centering
        \includegraphics[width=\linewidth]{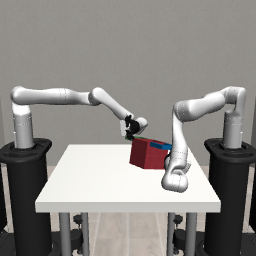}
        \caption{SR 24\% (F)}
    \end{subfigure}
    \hfill
    \begin{subfigure}[b]{0.19\linewidth}
        \centering
        \includegraphics[width=\linewidth]{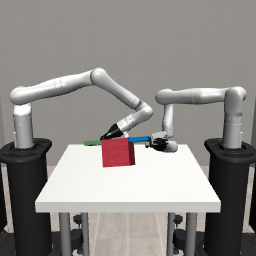}
        \caption{SR 48\% (S)}
    \end{subfigure}
    \hfill
    \begin{subfigure}[b]{0.19\linewidth}
        \centering
        \includegraphics[width=\linewidth]{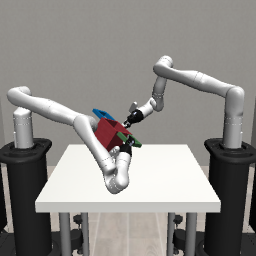}
        \caption{SR 52\% (S)}
    \end{subfigure}
    \makebox[\linewidth][c]{(a) Eureka}\par
    \vspace{5pt}

    \begin{subfigure}[b]{0.19\linewidth}
        \centering
        \includegraphics[width=\linewidth]{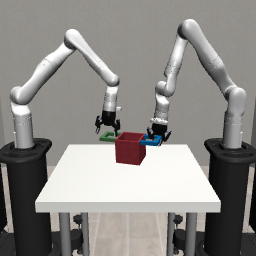}
        \caption{SR 0\% (F)}
    \end{subfigure}
    \hfill
    \begin{subfigure}[b]{0.19\linewidth}
        \centering
        \includegraphics[width=\linewidth]{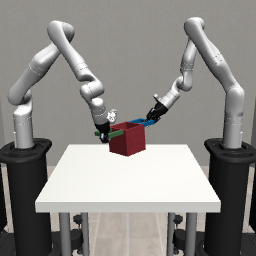}
        \caption{SR 80\% (S)}
    \end{subfigure}
    \hfill
    \begin{subfigure}[b]{0.19\linewidth}
        \centering
        \includegraphics[width=\linewidth]{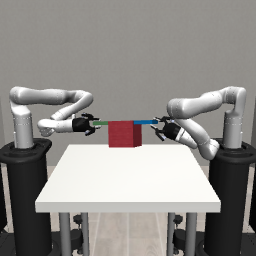}
        \caption{SR 88\% (S)}
    \end{subfigure}
    \hfill
    \begin{subfigure}[b]{0.19\linewidth}
        \centering
        \includegraphics[width=\linewidth]{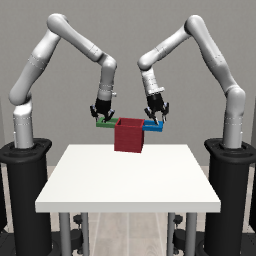}
        \caption{SR 96\% (S)}
    \end{subfigure}
    \hfill
    \begin{subfigure}[b]{0.19\linewidth}
        \centering
        \includegraphics[width=\linewidth]{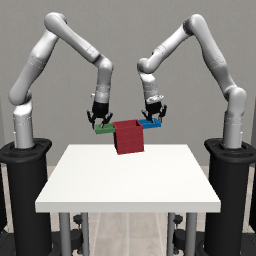}
        \caption{SR 100\% (S)}
    \end{subfigure}
    \makebox[\linewidth][c]{(b) CRAFT}\par
    \caption{Illustrative snapshots from the Lift task across $5$ trials for each method. Each snapshot is labeled with the policy success rate (SR) and whether the shown rollout succeeds (S) or fails (F). Eureka often achieves partial success but exhibits less physically feasible motions, whereas CRAFT produces more consistent successful rollouts. For Eureka, snapshots are taken from the first rollout of each trained policy. For CRAFT, snapshots are taken from the rollouts used by the policy evaluation module.}
    \label{fig:eureka_craft_full}
\end{figure}

In this section, we provide additional qualitative analysis of the experiments. As shown in Table~\ref{tab:success_rate}, Env Reward and Example Reward achieve low success rates across the tasks. Qualitatively, we observe that policies trained with Env Reward and Example Reward are susceptible to falling into local optima, resulting in suboptimal behaviors. For example, as shown in Fig.~\ref{fig:suboptimal_env_reward}, a policy trained with Env Reward converges to suboptimal strategies, such as grasping the pot but failing to lift it. This suggests that long-horizon tasks require reward components for multiple subtasks, such as reaching, grasping, and lifting, making it difficult to design a fixed reward whose global optimum consistently aligns with the desired behavior.

Additionally, as discussed in Section~\ref{subsec: results}, we observe that Eureka can produce behaviors that achieve partial task success but are difficult to transfer to real hardware. Fig.~\ref{fig:eureka_craft_full} shows representative rollouts from all $5$ trials of Eureka and CRAFT in Lift. Across Eureka trials, the learned policies often exhibit unnatural configurations to complete the task. We observe a common undesirable behavior in Eureka policies, where the arms twist in place near the pot instead of forming stable, physically plausible grasps. In the $8\%$ and $12\%$ success-rate policies, the robot often lifts the pot through incidental contact rather than a stable grasp. Even in the higher-performing $24\%$, $48\%$, and $52\%$ policies, the arms continue twisting around the pot and handles, including after partial grasping, instead of forming physically plausible grasps before lifting.

In contrast, although CRAFT has one failed trial with $0\%$ success due to unsuccessful handle grasping, its successful policies consistently reach more natural joint configurations. These behaviors are more suitable for hardware transfer, as the robots align with the handles and establish stable grasps before lifting. This comparison further emphasizes the benefit of staged curriculum learning: by progressively building intermediate skills, CRAFT guides policy learning toward more feasible coordination behaviors rather than optimizing only for final task success.

\begin{figure}
    \centering
    \begin{subfigure}[b]{0.4\linewidth}
        \centering
        \includegraphics[width=\linewidth]{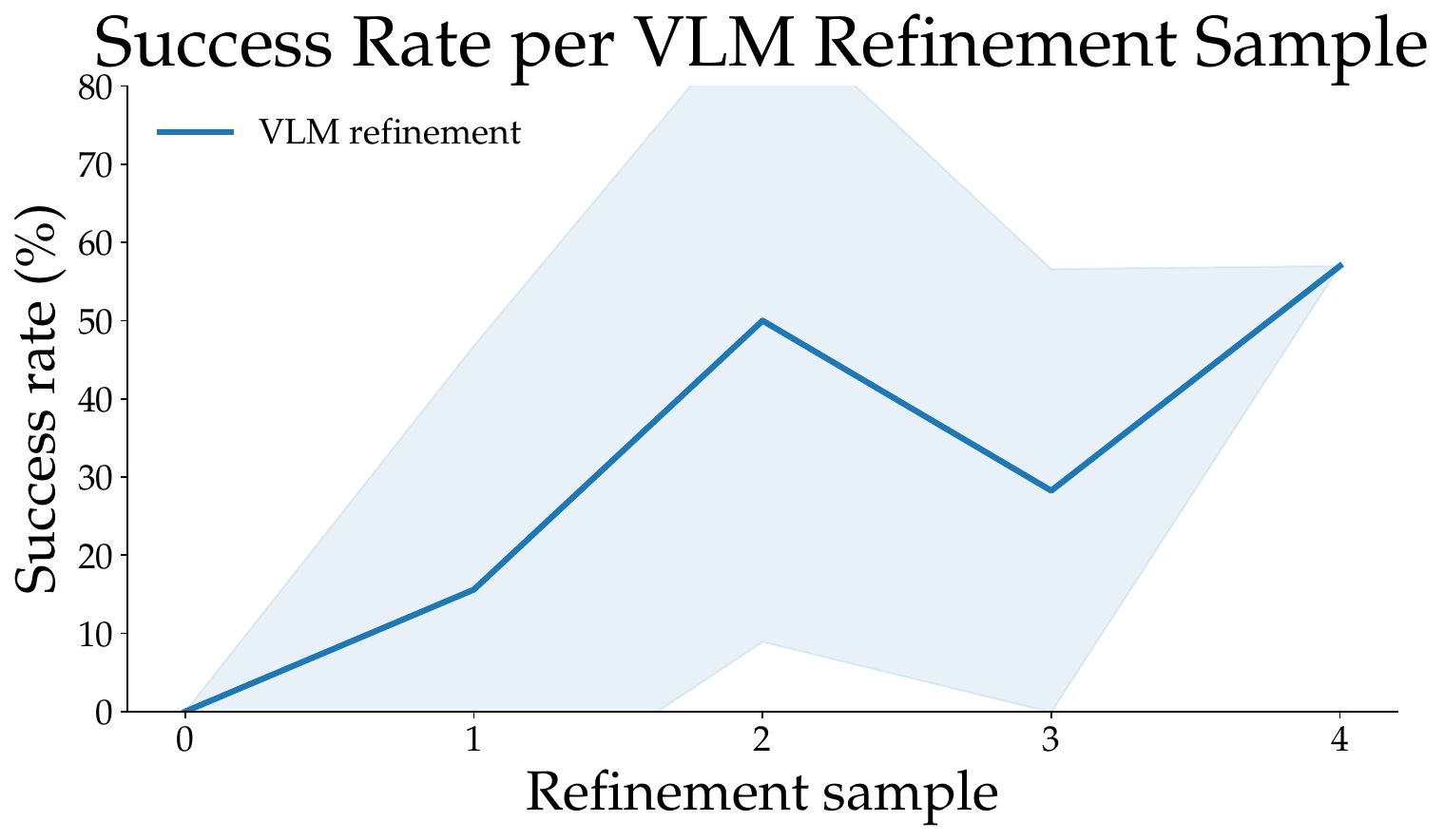}
        \caption{Success rate at each refinement iteration.}
        \label{fig:vlm_refinement_per_iter}
    \end{subfigure}
    \hfill
    \begin{subfigure}[b]{0.4\linewidth}
        \centering
        \includegraphics[width=\linewidth]{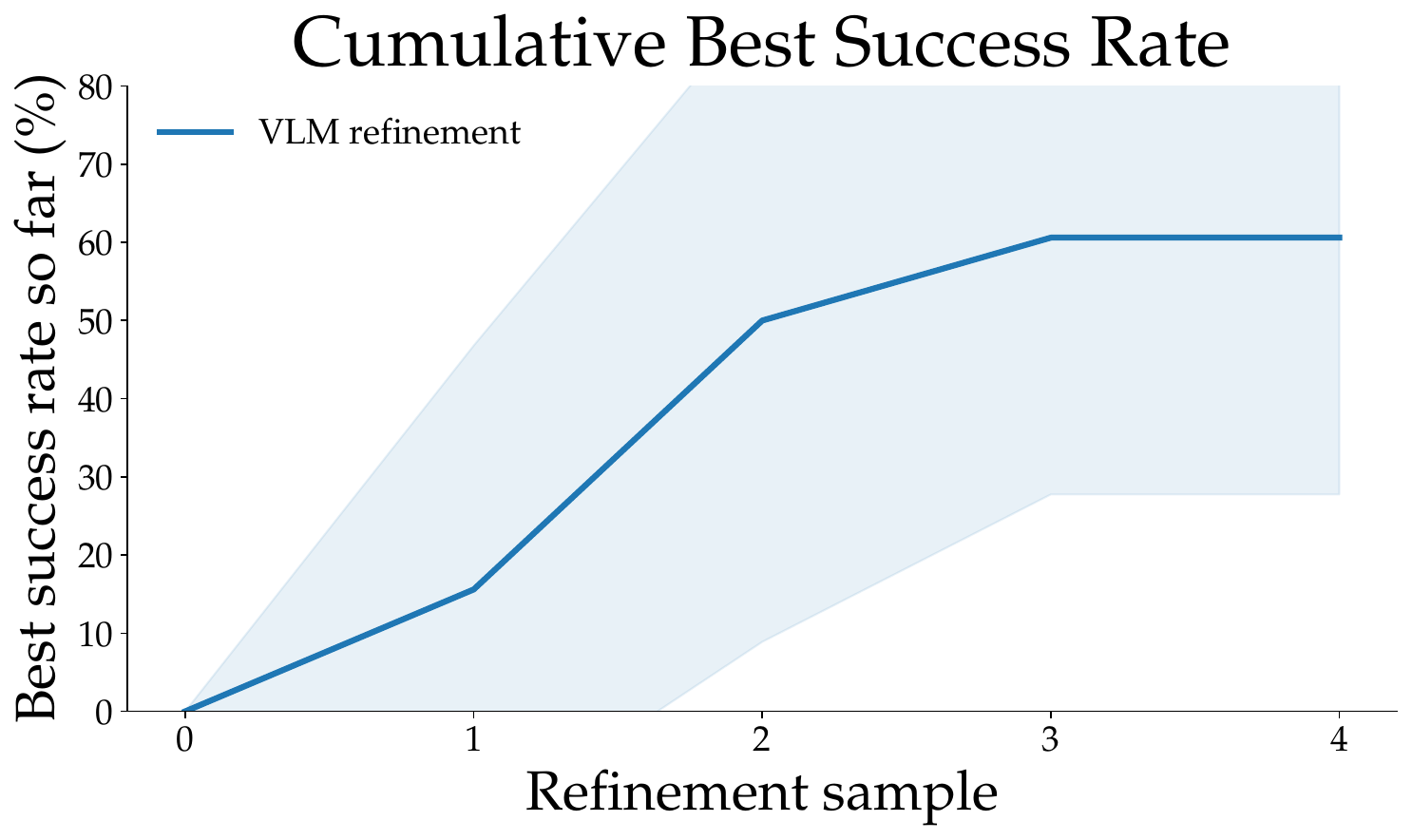}
        \caption{Best success rate up to each iteration.}
        \label{fig:vlm_refinement_best}
    \end{subfigure}
    \caption{Effect of VLM-guided reward refinement in the Gate environment without staged curriculum training. We report the mean and standard deviation over five runs. The per-iteration success rate measures the policy trained with the reward from each refinement iteration, while the cumulative best success rate reports the best performance achieved up to that iteration. Although performance drops at sample 3, it recovers at sample 4, showing that the refinement loop can overcome occasional unsuccessful updates.}
    \label{fig:vlm_refinement}
\end{figure}

To evaluate the efficiency of VLM-guided reward refinement in searching for effective reward functions, we conduct an additional experiment in the Gate environment. In this experiment, we directly apply VLM-guided reward refinement to the target task without staged curriculum training. Fig.~\ref{fig:vlm_refinement} reports the mean and standard deviation of the success rate over five runs across refinement iterations. After each refinement step, our VLM-guided reward refinement loop discovers improved reward functions, leading to policies with higher success rates. Starting from $0\%$ success, the method reaches an average success rate of approximately $60\%$ after five refinement iterations. These results demonstrate that VLM-guided refinement can efficiently improve reward functions even without curriculum-based staged training.

\subsection{Hardware Experiments} \label{subsec: hardware_exp}

\begin{figure}[h]
    \centering
    \begin{subfigure}[b]{0.32\linewidth}
        \centering
        \includegraphics[width=\linewidth]{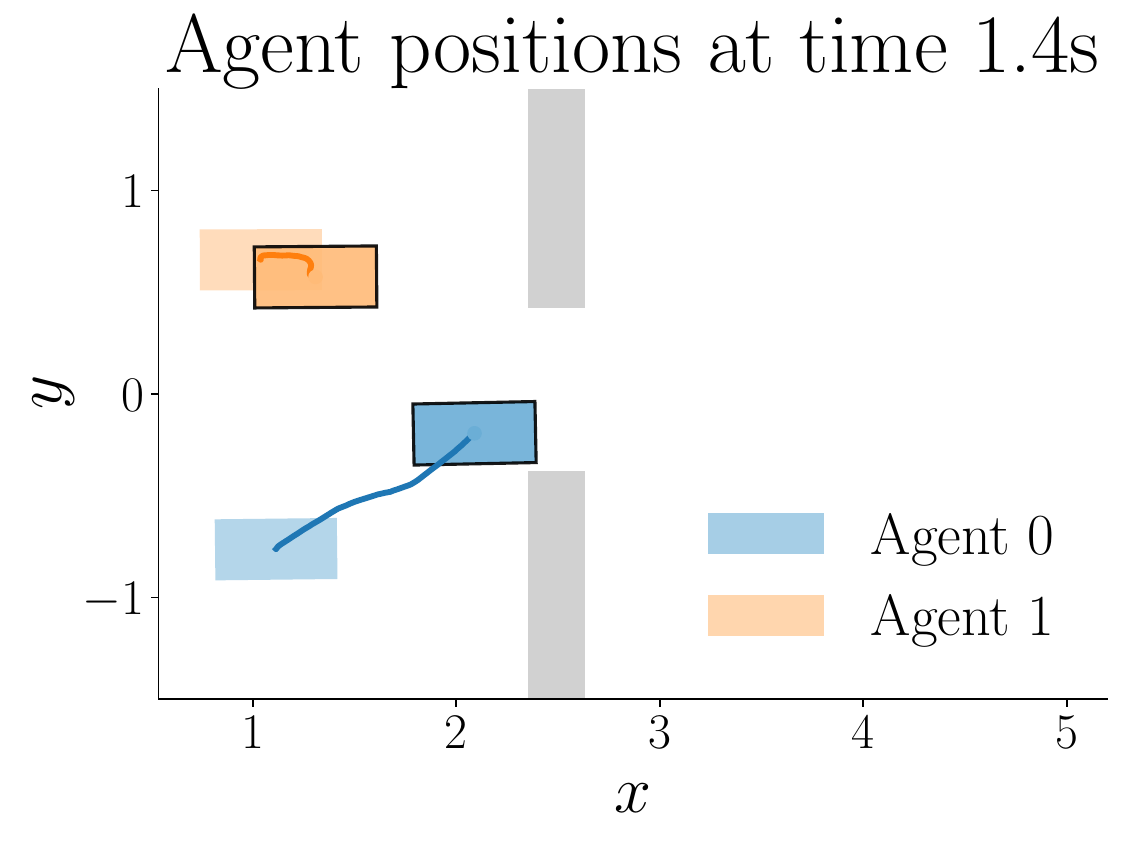}
    \end{subfigure}
    \hfill
    \begin{subfigure}[b]{0.32\linewidth}
        \centering
        \includegraphics[width=\linewidth]{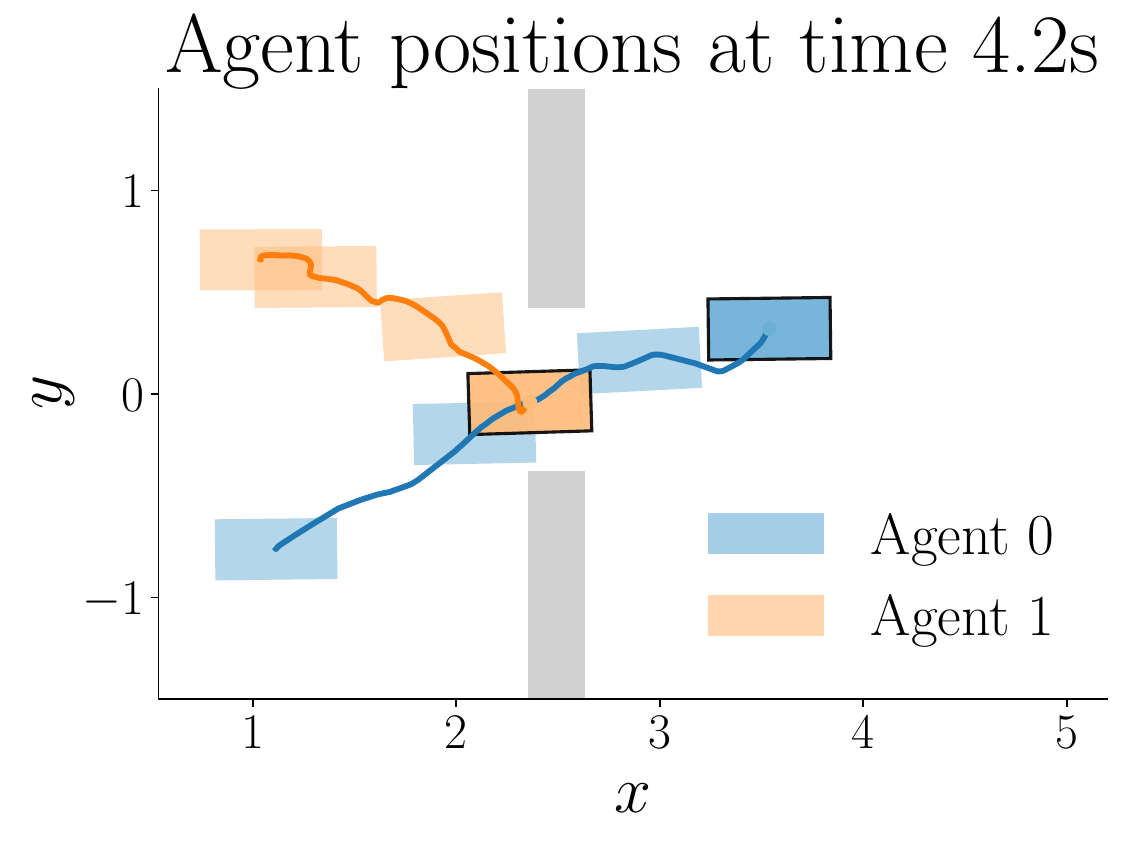}
    \end{subfigure}
    \hfill
    \begin{subfigure}[b]{0.32\linewidth}
        \centering
        \includegraphics[width=\linewidth]{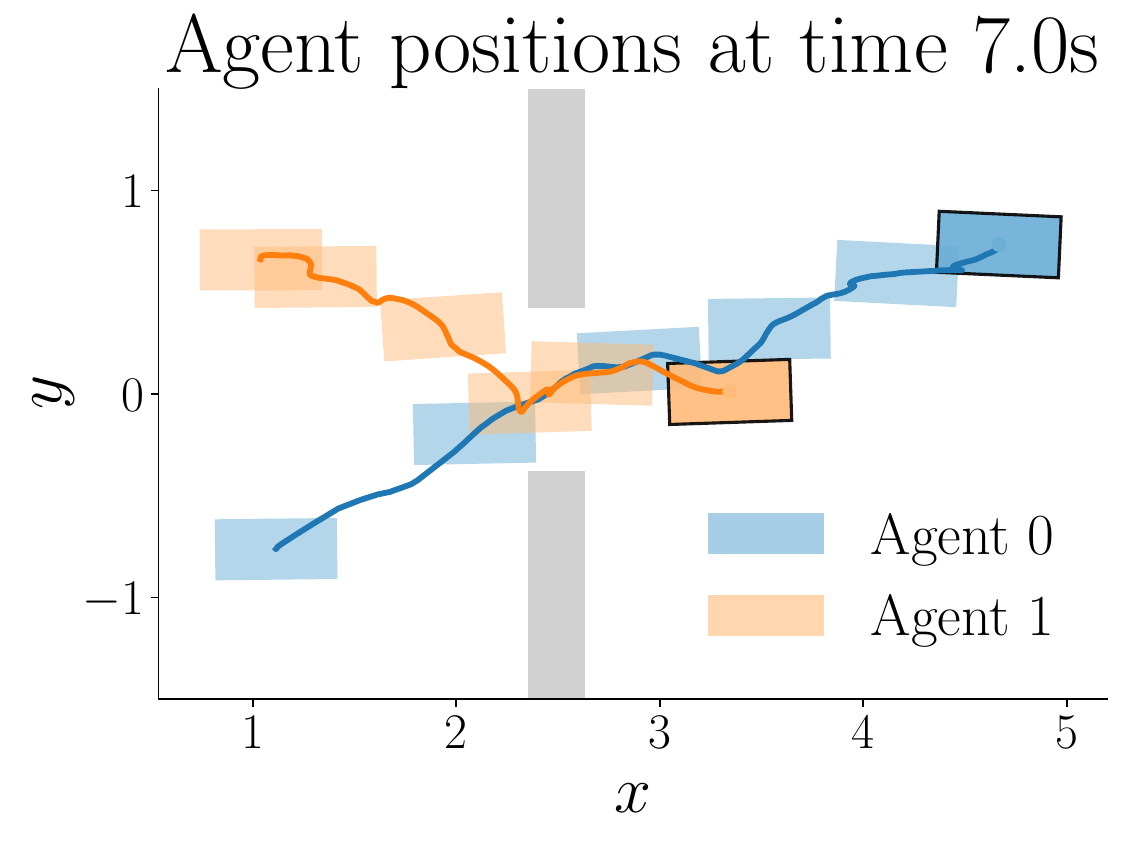}
    \end{subfigure}
    \caption{Trajectory of quadrupeds in our hardware experiments for Quadruped Gate policy trained with CRAFT. Two quadrupeds exhibit coordination behaviors in real world and manage to pass the gate sequentially without collision.}
    \label{fig:realwrod_traj}
\end{figure}

% For our hardware experiment of Gate, we use one Unitree Go2 and one Unitree Go1 robot. Since the embodiments and low-level MPC controllers differ from in simulation, we applied action component-wise scaling to the policy outputs before deployment. The initial pose of the quadrupeds and the pose of the gate is the same as in the simulation.
For hardware deployment, we used the same policy network parameters for both robots. The policy was trained in simulation using the Go2 model and was deployed without policy finetuning on one Unitree Go2 and one Unitree Go1. Since the Go1 and Go2 differ in embodiment and low-level MPC controllers, we applied a fixed robot-specific scaling to the velocity commands output by the policy before sending them to each robot. In particular, we found that the Go1 had difficulty moving forward when commanded velocities were small, so we used a larger scaling factor in the low-velocity regime for the Go1. This scaling was applied only as a post-processing step to the policy outputs. The policy parameters themselves were unchanged across robots.
Fig.~\ref{fig:realwrod_traj} shows example trajectory of Quadruped Gate environment in real-world hardware experiment. The two quadrupeds exhibit coordinated behavior: the blue robot passes through the gate first while the orange robot waits, then proceeds once the gate is clear.

\section{Example Prompts} \label{sec: prompt}
In this section, we provide prompts used in CRAFT. While the details may vary slightly across tasks, the structure remains the same. 
% Full prompts are available at \href{https://github.com/labicon/CRAFT}{github}.

\subsection{Curriculum Generation Module} 
We use two prompts which are given to the curriculum LLM. One for generating candidate curricula, and another for refining the final curriculum from the candidates.

\renewcommand{\lstlistingname}{Prompt}
\begin{lstlisting}[language=, caption=LLM prompt for generating curriculum candidates.]
You are a curriculum generator trying to generate a curriculum to solve multi-agent reinforcement learning tasks as effectively as possible. Your goal is to write a list of subtasks and corresponding reward function that will help agents to efficiently coordinate for target task.

Some helpful tips for generating a curriculum:
(1) Do not explore the world by doing random actions and do not try moving to manually defined position.
(2) Try to make the curriculum simple. Do not generate more than 5 tasks.
(3) The last task of your curriculum should align with the final goal of the environment.
(4) In curriculum, you cannot change the environment or the terminating condition. Focus on the task description.
(5) You are learning a centralized policy that have access to every agents.

You will be given several variables you can use to describe each subtask in curriculum.
Each task should be described using the given variables and do not introduce new variables.

Your output should be
Task 1
Name: 
Description: 
Reason: 

Task 2 
Name: 
Description: 
Reason: 

...

Fill out the names, descriptions, and reasons in the format.

<<Environment_Description>>

Variables you can use for curriculum descriptions are
<<State_Variables>>

These are the rules that you have to consider when writing a curriculum
(1) You cannot change start position of robots and objects.
(2) You cannot change the number of robots and objects.
(3) You should consider all robots in the environment.

\end{lstlisting}

\begin{lstlisting}[language=, caption=LLM prompt for refining curriculum candidates.]
You are a curriculum generator trying to generate a curriculum to solve multi-agent reinforcement learning tasks as effectively as possible. Your goal is to write a list of subtasks and corresponding reward function that will help agents to efficiently coordinate for target task.

You will be given three candidates of curriculums that were generated by other generators. 
Pick the best version and refine it if needed.
Note that the maximum possible curriculum is 5.

You will be given several variables you can use to describe each subtask in curriculum.
Each task should be described using the given variables and do not introduce new variables.

Your output should be
Task 1
Name: 
Description: 
Reason: 

Task 2 
Name: 
Description: 
Reason: 

...

Fill out the names, descriptions, and reasons in the format.

<<Environment_Description>>

Variables you can use for curriculum descriptions are
<<State_Variables>>

These are the rules that you have to consider when writing a curriculum
(1) You cannot change start position of robots and objects.
(2) You cannot change the number of robots and objects.
(3) You should consider all robots in the environment.

Three candidates of curriculums are:

Candidate 1:
<<Candidate_1>>

Candidate 2:
<<Candidate_2>>

Candidate 3:
<<Candidate_3>>
\end{lstlisting}

\subsection{Reward Function Generation Module} 

Below is the prompt given to the reward generation LLM for generating a base reward function.

\begin{lstlisting}[language=, caption=LLM prompt for generating base reward function.]
You should write reward function for given task using useful variables from the environment.
Your reward function is part of the curriculum learning which consists of learning sequence of different tasks.
You will be given description of past tasks, reward code for past tasks, and current task description.

Your gpt_reward(self) function's output should consist of three items
(1) the total reward, an integer
(2) a dictionary of each individual reward component where keys are names of reward component and items are each rewards
(3) the maximum possible reward that is used to normalize the reward, an integer

Some helpful tips for writing a reward function code:
(1) Use numpy or scipy functions to write the reward function. numpy, scipy is already imported as np and scipy. Do not import additional library.
(2) You may put higher weight on the reward for current task, but also include rewards from past tasks to avoid forgetting.
(3) Total reward must stay within [0, 1]. Always, scale the reward to have maximum 1. 
(4) Only use provided variables.
(5) Do not change other predefined parts in the code. 

<<Environment_Description>>

You can use help functions to calculate reward.
<<Helper_Functions>>

Example reward function for the task is 
<<Example_Reward>>

You should re-implement reward function to assign a proper reward function for given task. 
You can change the reward components that aligns well with the task.
Do not include any new inputs in the functions.

Note that you are designing a reward function for team, not a single agent. 

Generate a reward function code and command for
Task Name: <<Task_Name>>
Description: <<Task_Description>>
Reason: <<Task_Reason>>

Previous learned task was
Task Name: <<Task_Name>>
Description: <<Task_Description>>
Reason: <<Task_Reason>>
Code:
```python
<<Task_Code>>
```
\end{lstlisting}

\subsection{Policy Evaluation Module} 
Below is the prompt given to the evaluation VLM for evaluating a trained policy.

\begin{lstlisting}[language=, caption=VLM prompt for policy evaluation.]
You are a multi-agent reinforcement learning engineer asked to evaluate a rollout from a task in a curriculum. 

You will be provided with:
- Snapshot images from the rollout
- Trajectory data (state sequences)
- Previous and current task descriptions

Your job is to determine:
1. Whether the agents achieved the current goal.
2. Whether they have forgotten earlier subtasks.

Your output should be
Decision: [Success or Failure]

Reason: 
- [reason 1]
- [reason 2]
...

Fill out the [] brackets in the given format. (Don't keep the [] brackets)

Current task is:
<<Current task>>
Former tasks that RL succeeded to learn is:
<<Former tasks>>

Given images are snapshots in chronological order of the rollout.

Trajectory data:
<<Trajectory>>

Note that the distance and orientation is considered as ``very close'' if the value is smaller than 0.03 (This is NOT a threshold value, this is given to have a sense of how close they are)
Note that the trajectory data is sequential data for a whole episode.

Make your decision on both the trajectory data and the image.
\end{lstlisting}

\subsection{Reward Refinement Module} 
We use two prompts. One for generating advice on how to refine the reward (VLM), and one for refining the reward function given the advice (LLM).

\begin{lstlisting}[language=, caption=VLM prompt for generating reward refinement advice.]
You are a multi agent reinforcement learning engineer trying to refine a reward function for a given task.
You will be given: the reward function that was tried and failed, the experimental results, a description of why it failed, and the reward-component curve images.
First, analyze the reason why the reward function failed to learn the task.
Then, provide advice on how to improve or modify the reward function to achieve the task.

These are some tips in giving advice:
(1) Do not give more than three advice.
(2) You cannot add a penalty in the reward function, e.g. no reward can be a negative value.

These are the only functions that are allowed to used.
<<Helper_Functions>>
Information that cannot be obtained from the above functions cannot be used in the reward function.

Current task is:
<<Task_Name>>
<<Task_Description>>
<<Task_Reason>>

Reward function that was tried:
<<Reward_Function>>

Failure reason:
<<Failure_Reason>>

Do not give an advice on the curriculum of the reward. Just consider how to improve the reward function given the current task.
Do not provide the whole code of the reward function, just give advice on how to improve it.
\end{lstlisting}

\begin{lstlisting}[language=, caption=LLM prompt for refining a reward function using advice.]
You are a multi agent reinforcement learning engineer trying to refine a reward function for the given task.
The reward function is part of the curriculum learning which consists of learning sequence of different tasks.
You will be given description of past tasks, a reward function that was tried and failed, and an advice on how to improve it.
Follow the advice and provide a new reward function that should be able to achieve the given task.
Do not repeat the advice or the old reward function in your answer.

Your gpt_reward(self) function's output should consist of two items
    (1) the total reward, an integer
    (2) a dictionary of each individual reward component where keys are names of reward component and items are each rewards
    (3) the maximum possible reward that is used to normalize the reward, an integer

Some helpful tips for writing a reward function code:
(1) Use numpy or scipy functions to write the reward function. numpy, scipy is already imported as np and scipy. Do not import additional library.
(2) You may put higher weight on the reward for current task, but also include rewards from past tasks to avoid forgetting.
(3) Total reward must stay within [0, 1]. Always, scale the reward to have maximum 1. 
(4) Only use provided variables.
(5) Do not change other predefined parts in the code. 

You can use help functions to calculate reward.
<<Helper_Functions>>

Current task is:
<<Current_Task>>
Former tasks that RL succeeded to learn is:
<<Former_Tasks>>

Reward function that was tried:
<<Reward_Function>>

Advice on how to improve the reward function:
<<Advice>>

Do not deviate from the reward structure given in the example.
Note that the advice might not acount for what information it can use in the reward function. So not all the advice might be applicable.
\end{lstlisting}

\subsection{Task-Specific Substitutions for Two Arm Lift} 
Below are task-specific substitutions (e.g., environment details and task descriptions) that are inserted into the general prompts above.

\begin{lstlisting}[language=, caption=\textless\textless Environment\_Description\textgreater\textgreater for Two Arm Lift.]
A large pot with two handles is placed on a table top. The pot is rectangular shaped with U-shaped handles on the opposite side of the pot rim. The grip of the handle is level, parallel to the tabletop. Two robot arms are placed on opposite ends of the table. The two robot arms must each grab a handle and lift the pot together, above a certain height (0.1m), while keeping the pot level (no more tilted than 30 degrees). The pot location is randomized at the beginning of each episode.
\end{lstlisting}

\begin{lstlisting}[language=, caption=\textless\textless State\_Variables\textgreater\textgreater for Two Arm Lift.]
- Euclidean distance from each robot's end-effector and its target handle
- Orientation difference between each robot's end-effector and its target handle
- Gripper state of each robot (whether it grasped the handle or not)
- Pot elevation above the table
- Pot tilt as cosine of the angle between its normal and the vertical
\end{lstlisting}

\begin{lstlisting}[language=, caption=\textless\textless Helper\_Functions\textgreater\textgreater for Two Arm Lift.]
self._get_check_grasps(): returns two boolean value for each arm whether or not it's grasping the handle or not.
self._get_pot_elevation(): returns the distance (in meters) from pot's bottom to the table top
self._get_tilt_degree(): returns the cosine value of the tilted degree of the pot. The tilted degree is the difference between the z axis and pot's normal vector
self._get_gripper_to_handle_distance(): returns two float value of distance from each robot's end-effector to each handle
self._get_gripper_to_handle_rotation(): returns two rotation difference (calculated with R1.T @ R2) between each robot's end-effector to each handle with shape: (3,3) each
self._check_success(): returns True if the pot is higher than 0.1
\end{lstlisting}

\renewcommand{\lstlistingname}{Code}
\setcounter{lstlisting}{0}
\begin{lstlisting}[language=Python, caption=\textless\textless Example\_Reward\textgreater\textgreater for Two Arm Lift.]
def gpt_reward(self):

    self.gpt_rew = 0
    self.gpt_rew_dict = {}
    max_reward = 7.5

    # reward components
    reach_reward = 0.0
    align_reward = 0.0
    lift_reward = 0.0
    grasp_reward = 0.0
    success_reward = 0.0

    # check if pot is tilted less than 30 degrees
    cos_z = self._get_tilt_degree()
    cos_30 = np.cos(np.pi / 6)
    direction_coef = 1 if cos_z >= cos_30 else 0

    # check for goal completion: cube is higher than the table top above a margin
    if self._check_success():
        success_reward += max_reward * direction_coef

    else:

        # Reaching reward - max 0.2
        _g0h_dist, _g1h_dist = self._get_gripper_to_handle_distance()
        reach_reward += 0.1 * (1 - np.tanh(10.0 * _g0h_dist))
        reach_reward += 0.1 * (1 - np.tanh(10.0 * _g1h_dist))

        # aligning reward - max 0.3
        _g0h_rot, _g1h_rot = self._get_gripper_to_handle_rotation()
        _g0h_rot_diff = np.trace(np.eye(3) - _g0h_rot) 
        _g1h_rot_diff = np.trace(np.eye(3) - _g1h_rot) 
        align_reward += 0.15 * (1 - np.tanh(10.0 * _g0h_rot_diff))
        align_reward += 0.15 * (1 - np.tanh(10.0 * _g1h_rot_diff))

        # lifting reward - max 5.0
        elevation = self._get_pot_elevation()
        r_lift = min(max(elevation, 0), 0.2)
        lift_reward += 25.0 * direction_coef * r_lift

        # Grasping reward - max 2.0
        g0_grasp, g1_grasp = self._get_check_grasps()
        if g0_grasp:
            grasp_reward += 1.0
        if g1_grasp:
            grasp_reward += 1.0

    # Add and normalize the reward
    self.gpt_rew = reach_reward + align_reward + lift_reward + grasp_reward + success_reward
    self.gpt_rew *= 1.0 / max_reward

    self.gpt_rew_dict = {
        'reach_reward': reach_reward,
        'align_reward': align_reward,
        'lift_reward': lift_reward,
        'grasp_reward': grasp_reward,
        'success_reward': success_reward,
    }
    
    return self.gpt_rew, self.gpt_rew_dict, max_reward
\end{lstlisting}

\section{Examples in VLM-guided Reward Refinement}

This section provides example results of VLM-guided reward refinement in quadruped seesaw environment.

\subsection{Example Results of Policy Evaluation}

\begin{figure}[h]
    \centering
    \includegraphics[width=0.18\linewidth]{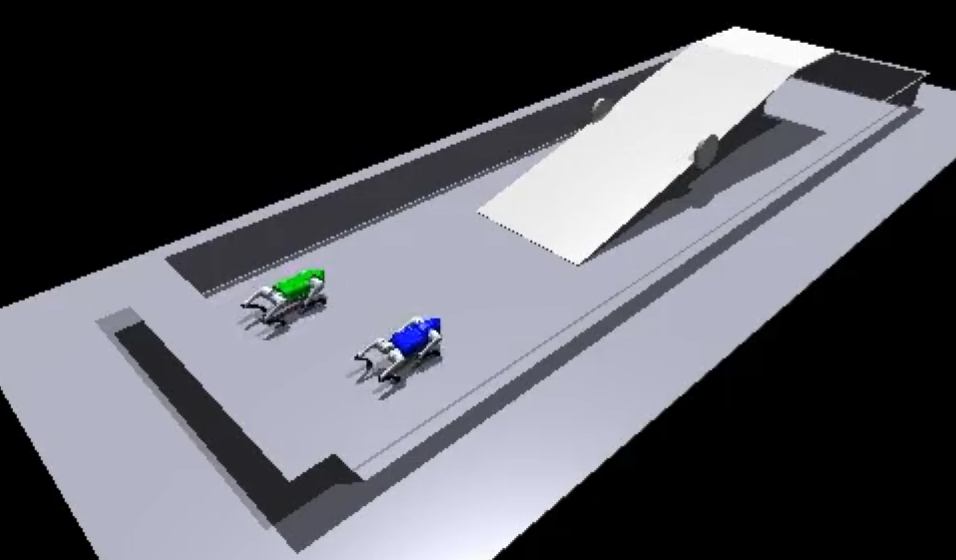}
    \includegraphics[width=0.18\linewidth]{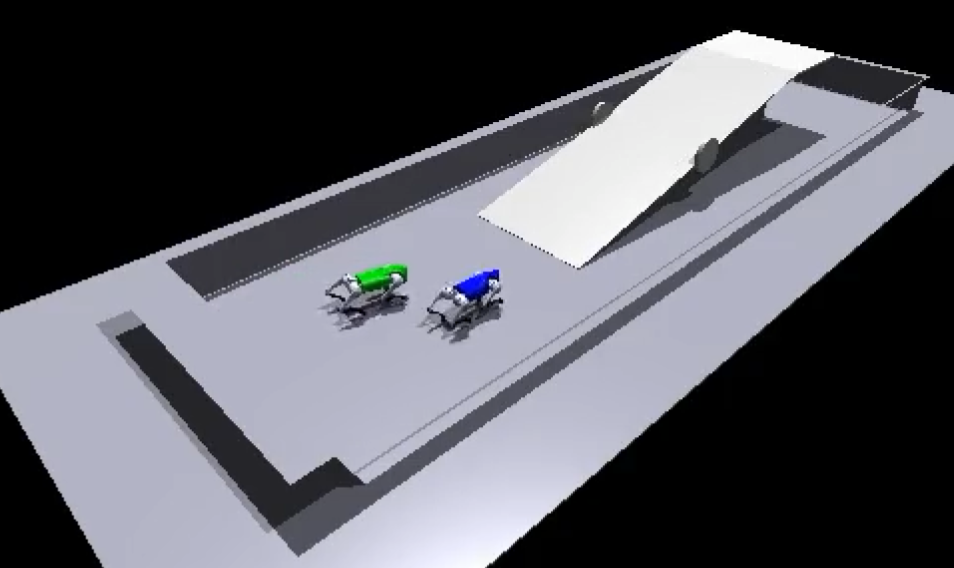}
    \includegraphics[width=0.18\linewidth]{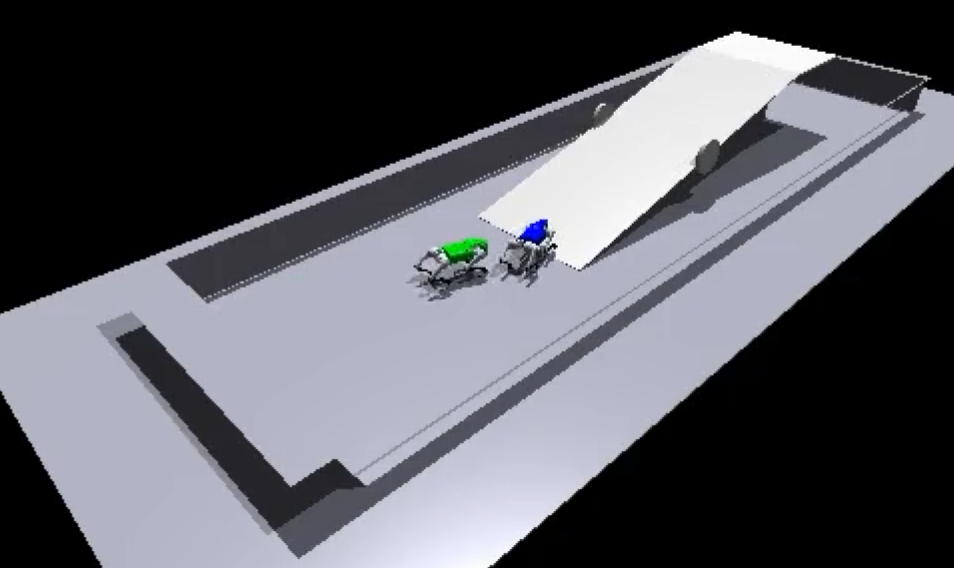}
    \includegraphics[width=0.18\linewidth]{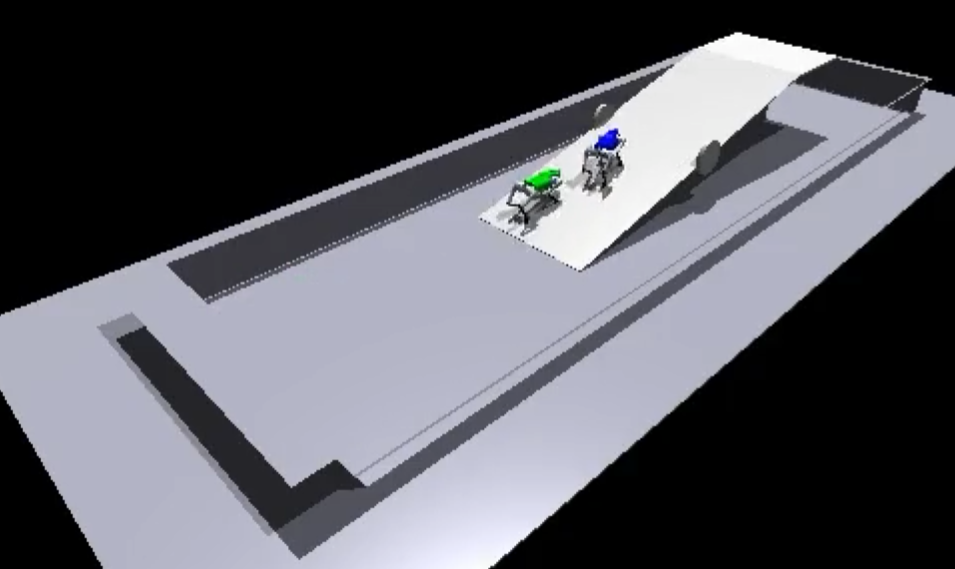}
    \includegraphics[width=0.18\linewidth]{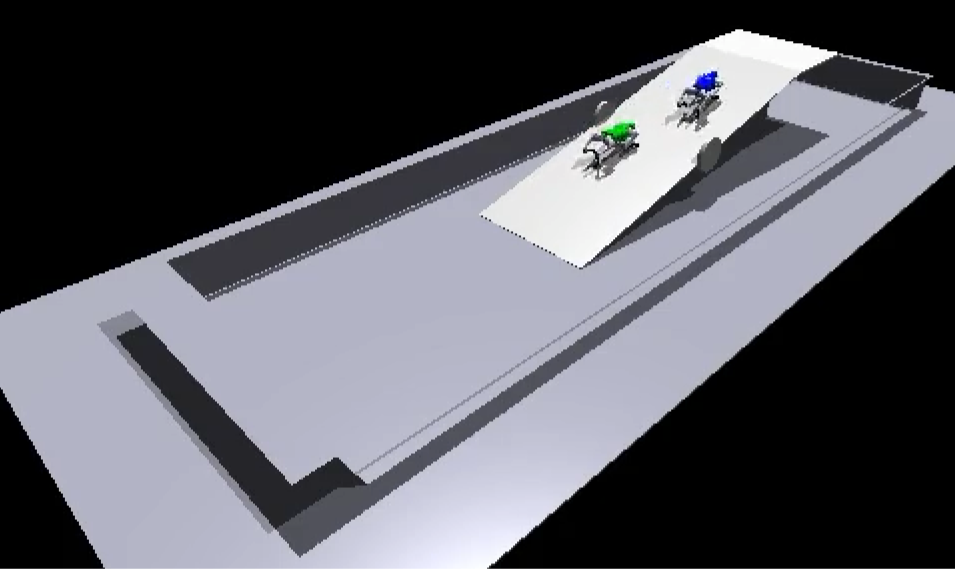}
    \caption{Rollout snapshots provided to the policy evaluation VLM.}
    \label{fig:eval_rollout}
\end{figure}

\renewcommand{\lstlistingname}{Prompt}
\setcounter{lstlisting}{9}
\begin{lstlisting}[language=, caption=Example VLM prompt for evaluating `seesaw entry and stability task' in Quadruped Seesaw.]
Current task is:
Task Name: 2_Seesaw Entry and Stability Task
Description: Train both agents to coordinate so that Agent 0 enters and traverses the seesaw toward the end at position [x=7, y=0], ensuring gradual and controlled movement to maintain balance. Meanwhile, Agent 1 enters the seesaw afterward, stopping and stabilizing at the start position [x=4, y=0, z=1.0].
Reason: Establishing sequential entry and motion ensures Agent 0 can proceed unimpeded while Agent 1 stabilizes the seesaw. Effective coordination during this phase prevents excessive tilting, allowing for smooth advancement toward the goal.
Former tasks that RL succeeded to learn is:

Task Name: 1_Initial Alignment Task
Description: Ensure Agent 0 (Blue) aligns itself with the start of the seesaw at position [x=4, y=0, z=1.0] while Agent 1 (Green) maintains its initial position, preparing to follow Agent 0. Both agents should aim to keep y position near 0 to stay within the seesaw's width.
Reason: This task sets the foundational entry sequence where Agent 0 leads the way, ensuring it is ready to proceed onto the seesaw without obstruction from Agent 1. This preparation is crucial for establishing entry control and ensuring proper seesaw engagement.

Given images are snapshots in chronological order of the rollout.
Agent 0 is colored with blue and Agent 1 is colored with green.

Trajectory data:
seesaw_center:
[5.500 0.000 1.000]
seesaw_start:
[4.000 0.000]
seesaw_end:
[7.000 0.000]
target_pos:
[7.700 0.000 1.500]
agent_0_pos:
[[0.920 -0.730 0.420]
 [1.380 -0.550 0.270]
 [2.130 -0.250 0.280]
 [2.910 -0.070 0.290]
 [3.640 -0.010 0.330]
 [4.360 0.030 0.520]
 [5.210 0.140 0.740]
 [5.910 0.320 0.920]
 [6.250 0.240 0.960]
 [6.910 0.210 1.150]
 [7.280 0.200 1.110]
 [7.250 0.080 0.970]
 [7.200 -0.030 1.010]
 [7.270 -0.020 1.240]
 [7.850 0.110 1.300]
 [7.950 -0.080 1.290]
 [7.980 -0.250 1.280]
 [8.070 -0.370 1.280]
 [8.160 -0.450 1.220]
 [8.160 -0.390 1.220]
 [8.090 -0.370 1.220]]
agent_1_pos:
[[0.920 0.830 0.420]
 [1.120 0.850 0.270]
 [1.500 0.600 0.280]
 [2.190 0.400 0.290]
 [2.890 0.370 0.280]
 [3.590 0.390 0.350]
 [4.140 0.390 0.450]
 [4.210 0.370 0.430]
 [4.150 0.180 0.420]
 [4.210 0.160 0.450]
 [4.320 0.200 0.630]
 [4.380 0.150 0.740]
 [4.380 0.260 0.610]
 [4.380 0.120 0.460]
 [4.460 -0.050 0.510]
 [4.510 0.070 0.570]
 [4.400 0.040 0.530]
 [4.440 0.070 0.550]
 [4.340 -0.010 0.510]
 [4.300 0.010 0.470]
 [4.220 -0.110 0.420]]

Note that this is not the final goal in the curriculum, don't be too strict on the decision.
\end{lstlisting}

\begin{lstlisting}[language=, caption=Example VLM answer for evaluating `seesaw entry and stability task' in Quadruped Seesaw.]
Decision: Success

Reason:
- Agent 0 led the sequence, entered the seesaw first, and traversed to and beyond the target end (around x = 7.2-8.2) with smooth, controlled height (z) changes, indicating balance was maintained.
- Agent 1 only stepped onto the seesaw after Agent 0 had committed, then remained stably near the start region (x ~ 4.2, y ~ 0, small z-oscillations), satisfying the stabilization requirement.
- The initial-alignment behaviour is intact (Agent 1 kept close to y = 0 while Agent 0 aligned to the seesaw before entry), showing no forgetting of the previous subtask.
\end{lstlisting}

\subsection{Example Results of Reward Advice}

\begin{figure}[h]
    \centering
    \includegraphics[width=0.22\linewidth]{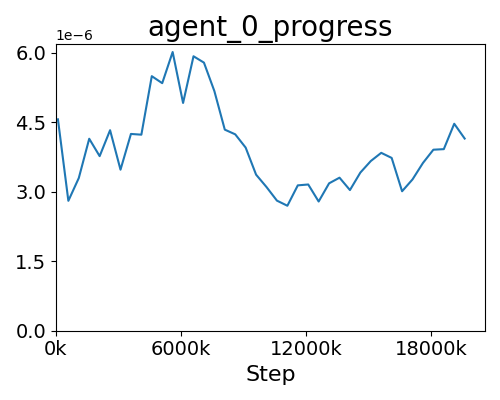}
    \includegraphics[width=0.22\linewidth]{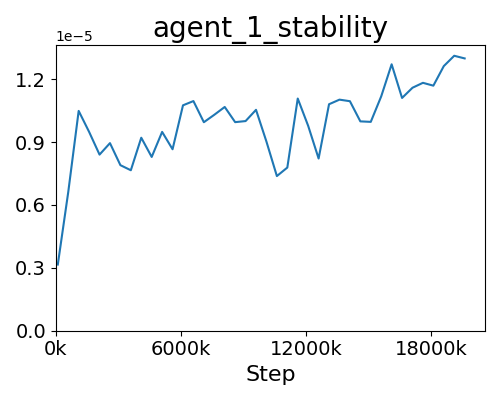}
    \includegraphics[width=0.22\linewidth]{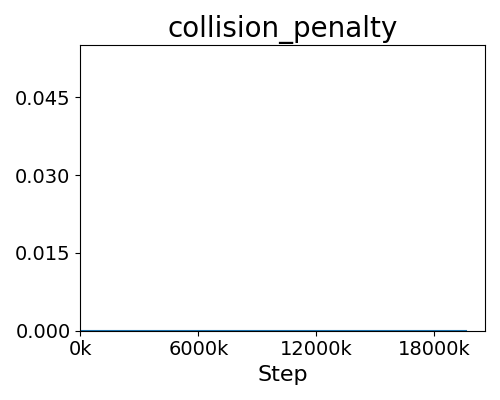}
    \includegraphics[width=0.22\linewidth]{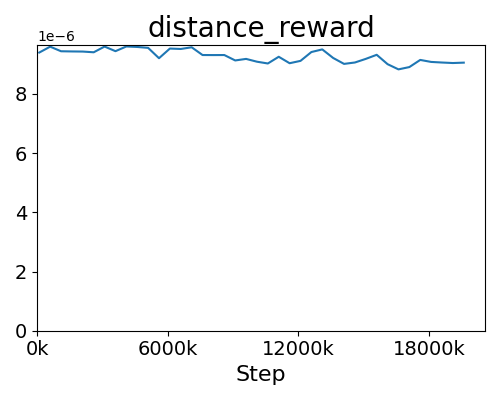} \\[0.5em]
    \includegraphics[width=0.22\linewidth]{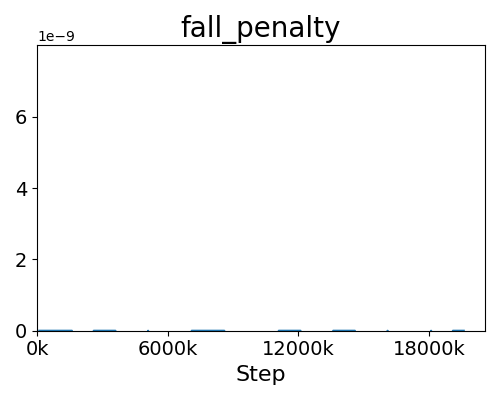}
    \includegraphics[width=0.22\linewidth]{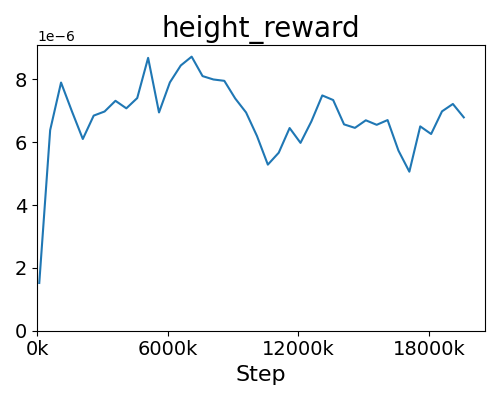}
    \includegraphics[width=0.22\linewidth]{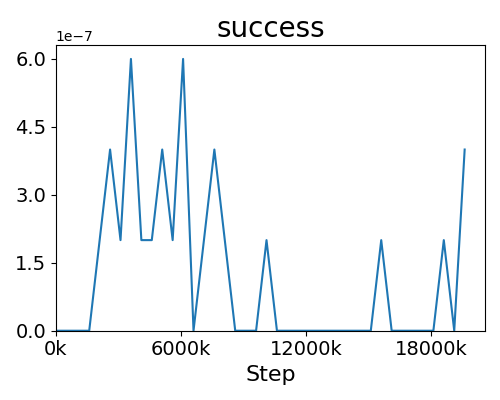}
    \includegraphics[width=0.22\linewidth]{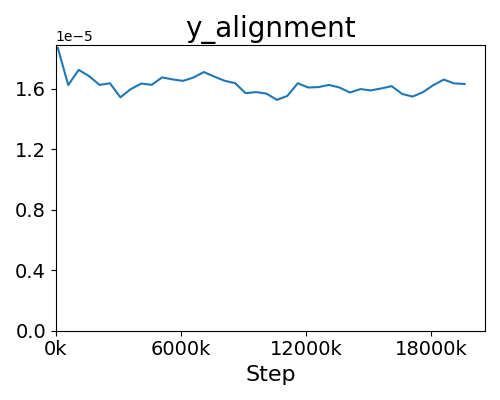}
    \caption{Reward component learning curves provided to the reward advice VLM.}
    \label{fig:learning_curve}
\end{figure}

\begin{lstlisting}[language=, caption=Example VLM prompt for evaluating `target platform ascension task' in Quadruped Seesaw.]
Current task is:
Name: Target Platform Ascension Task  
Description: Ensure Agent 0 continues confidently toward the end of the seesaw and reaches the target platform [x coordinate > 7, height > 1.3], while Agent 1 maintains stability at the seesaw's starting point. Agent 1 should ensure it provides enough counterbalance to prevent the seesaw from destabilizing.  
Reason: This task focuses on culmination and assessment of learned coordination strategies, where Agent 0 must finalize its movement to reach the target while relying on Agent 1 to maintain seesaw balance. Successfully reaching the platform signifies the completion of the objective and tests the agents' ability to coordinate effectively under the primary task constraints.

Reward function that was tried:
def gpt_reward(self, state, action):

    reward = torch.zeros((self.num_envs, self.num_agents), device=self.env.device)
    rew_dict = {}
    max_reward = 23.0

    # Reward agent 0 for progressing towards the target platform - max 3.0
    agent_0_progress_reward = 3.0 * self._progress_to_target(state, action)
    # Only reward agent 0 when progressing towards the target
    agent_0_progress_reward[:, 1] = 0.0
    agent_0_progress_reward = self._check_reward_shape(agent_0_progress_reward)
    reward += agent_0_progress_reward
    rew_dict['agent_0_progress'] = torch.mean(agent_0_progress_reward[:, 0])

    # Reward agent 1 for holding position at the seesaw start - max 3.0
    agent_1_stability_reward = 3.0 * self._progress_to_seesaw_start(state, action)
    # Only reward agent 1 for maintaining stability
    agent_1_stability_reward[:, 0] = 0.0
    agent_1_stability_reward = self._check_reward_shape(agent_1_stability_reward)
    reward += agent_1_stability_reward
    rew_dict['agent_1_stability'] = torch.mean(agent_1_stability_reward[:, 1])

    # Reward for maintaining y position close to 0 for both agents - max 2.0
    y_alignment_reward = 2.0 * self._y_alignment(state, action)
    y_alignment_reward = self._check_reward_shape(y_alignment_reward)
    reward += y_alignment_reward
    rew_dict['y_alignment'] = torch.mean(y_alignment_reward)

    # Reward for height progression for Agent 0, aiming to reach above 1.3m - max 2.0
    height_reward = 2.0 * self._normalized_height(state, action)
    height_reward[:, 1] = 0.0  # Only agent 0 should get height reward
    height_reward = self._check_reward_shape(height_reward)
    reward += height_reward
    rew_dict['height_reward'] = torch.mean(height_reward[:, 0])

    # Penalty for falling off the seesaw - min -2.0
    fall_penalty = -2.0 * self._fall(state, action).float()
    fall_penalty = self._check_reward_shape(fall_penalty)
    reward += fall_penalty
    rew_dict['fall_penalty'] = torch.mean(fall_penalty)

    # Penalty for colliding with walls or structures - min -1.0
    collision_penalty = -1.0 * self._wall_collision(state, action).float()
    collision_penalty = self._check_reward_shape(collision_penalty)
    reward += collision_penalty
    rew_dict['collision_penalty'] = torch.mean(collision_penalty)

    # Reward for maintaining an appropriate distance between agents - max 1.0
    agent_distance = self._agent_distance(state, action)
    distance_reward = 1.0 * torch.clamp(torch.square(agent_distance), max=1.0)
    distance_reward = self._check_reward_shape(distance_reward)
    reward += distance_reward
    rew_dict['distance_reward'] = torch.mean(distance_reward)

    # Reward for success when Agent 0 reaches the platform - max 10.0
    success = self._success(state, action)
    success_reward = 10.0 * success.float()
    success_reward = self._check_reward_shape(success_reward)
    reward += success_reward
    rew_dict['success'] = torch.mean(success_reward)

    # Normalize the reward
    reward = self._check_reward_shape(reward)
    reward /= max_reward

    return reward, rew_dict, max_reward

Failure reason:
- Agent 0 never reached the target platform region defined by x>7 and height z>1.3; its maximum z was only ~1.1 near x~6.86 before falling back down.  
- Agent 1 did not hold the seesaw at the start point (x~4); it remained around x=0.8~0.9 throughout, indicating it forgot the earlier alignment and stability subtasks.

Do not give an advice on the curriculum of the reward. Just consider how to improve the reward function given the current task.
Do not provide the whold code of the reward function, just give advices on how to improve it.
\end{lstlisting}

\begin{lstlisting}[language=, caption=Example VLM advice for improving reward function for `target platform ascension task' in Quadruped Seesaw.]
Here are three focused suggestions for reshaping your reward so that Agent 0 is driven all the way up onto the platform and Agent 1 stays firmly at the seesaw pivot:

1.  Re-weight and de-crowd your signal for Agent 0's forward/vertical progress  
   - Right now ``agent_0_progress'' (*3) and ``height_reward'' (*2) get drowned in the many small terms and then all get divided by 23.  Even at full scale they only amount to ~0.1 per step.  Boost those two terms by an order of magnitude (e.g. *20-30 for progress, *10-15 for height) and then remove or dramatically shrink the less critical terms (y_alignment, distance_reward) so that the agent sees a strong, unambiguous gradient pulling it to x>7 and height>1.3.  

2.  Turn your fall-and-collision ``penalties'' into positive stay-alive/balance bonuses  
   - You've defined negative fall/collision terms, but the spec forbids negative rewards anyway, and they barely ever fire (1e-9 scale).  Instead give a small constant ``survival'' bonus every step that you remain upright and collision-free (e.g. +1 per step of life/collision-free).  This keeps all components >0 and provides a steady shaping signal without explicit negatives.

3.  Concentrate Agent 1's reward on seesaw balance, not just position drift  
   - '_progress_to_seesaw_start' only measures how close Agent 1 got initially, but says nothing about ongoing balance. Repurpose your y_alignment term (or distance_reward) so that it's applied only to Agent 1 and heavily weighted: reward 1 - tilt (i.e. how flat the seesaw is) or reward the inverse of Agent 1's deviation from the pivot. This gives Agent 1 a clear continuous incentive to lock itself at the fulcrum and keep the seesaw horizontal while Agent 0 climbs.
\end{lstlisting}